\documentclass{article}

\makeatletter
\def\input@path{{template/}}
\makeatother

\usepackage{iclr2027_conference,times}

\iclrfinalcopy 

\usepackage{amsmath,amsfonts,bm}

\def\eqref#1{equation~\ref{#1}}

\def\1{\bm{1}}

\DeclareMathAlphabet{\mathsfit}{\encodingdefault}{\sfdefault}{m}{sl}
\SetMathAlphabet{\mathsfit}{bold}{\encodingdefault}{\sfdefault}{bx}{n}

\usepackage[hidelinks]{hyperref}
\usepackage{url}
\usepackage{array}
\usepackage{cleveref}

\usepackage{booktabs}
\usepackage{multirow}
\usepackage{longtable}
\usepackage{placeins}
\usepackage{amsmath,amssymb}
\usepackage{algorithm}
\usepackage{algorithmic}
\usepackage{graphicx}
\usepackage{subcaption}
\usepackage{wrapfig}
\usepackage{enumitem}
\usepackage{adjustbox}
\usepackage{tikz}
\usetikzlibrary{arrows.meta,positioning,calc,fit,backgrounds,decorations.markings}
\usepackage{xcolor}
\definecolor{pactaccent}{RGB}{28,96,150}
\definecolor{revblue}{RGB}{0,70,200}
\definecolor{revred}{RGB}{200,0,0}

\renewcommand{\eqref}[1]{(\ref{#1})}

\usepackage[normalem]{ulem}

\title{PACT: Plan-Augmented Coordination via \\
Trace-Graphs for Dynamic Multi-Agent\\ Combinatorial Optimization}

\author{Xiao Zhou, Alex Schutz, Victor-Alexandru Darvariu \& Nick Hawes \\
Oxford Robotics Institute\\
University of Oxford\\
OX1 3PJ, Oxford, United Kingdom. \\
\texttt{\{xiaozhou,alexschutz,victord,nickh\}@robots.ox.ac.uk} 
}

\begin{document}
\maketitle

\begin{abstract}
Multi-agent approaches have been investigated recently as a means of improving the scalability of reinforcement learning for neural combinatorial optimization.
Existing methods assume that all problem instance information is known upfront. However, real-world problems are often \textit{dynamic}: relevant information is progressively revealed as the agents take actions (e.g., arriving at an edge in a routing problem only to discover it is untraversable). 
In such cases, existing methods fail to coordinate agents to explore and exploit the progressively revealed information.
To address the arising coordination challenge, we introduce PACT (\textbf{P}lan-\textbf{A}ugmented \textbf{C}oordination via \textbf{T}race-graphs).
PACT augments the environment graph with auxiliary nodes representing the plans of agents, which are connected to the resource nodes to indicate future dependencies. 
The proposed message-passing scheme over the augmented graph grounds inter-agent coordination in the environment and enables proactive, rather than reactive, conflict resolution.
We show that PACT improves solution quality in dynamic problems relative to baselines and exhibits the best transfer to larger instances on which, remarkably, it outperforms centralized oracles with privileged access to instance uncertainty.

\end{abstract}

\begingroup
\captionsetup{skip=2pt}
\captionsetup[figure]{skip=4pt}
\section{Introduction}
\label{sec:intro}

Many routing, scheduling, and allocation problems require several agents to construct a joint combinatorial solution under a shared objective such as total travel cost or makespan, a setting known as multi-agent combinatorial optimization (MACO) \citep{berto2025parco}. Existing formulations typically assume that the instance to be solved is fully specified before construction begins. In many real systems, however, relevant information becomes available only through execution: a route may be discovered to be untraversable only when an agent reaches it, or a machine may become known to be unavailable only when an operation is attempted \citep{pillac2013review, ouelhadj2009survey}. We call this setting \textbf{dynamic multi-agent combinatorial optimization} (DMACO).

Classic combinatorial optimization typically assumes fully disclosed instances upfront. Although reactive re-optimization can adapt to newly revealed information, solving must recur continuously \citep{vieira2003rescheduling,bertsimas2019online}, and such reactive paradigms generally do not account for how current decisions affect information and coordination requirements that arise later \citep{ulmer2019anticipation}. Neural combinatorial optimization (NCO) amortizes the cost of re-solving by learning policies over instance distributions \citep{kool2019attention, kwon2020pomo, kool2022dpdp} and has recently been extended to multi-agent settings \citep{berto2025parco, luttmann2026macsim}. Nonetheless, these NCO approaches still assume fixed and fully observable instances prior to execution.

Since information in DMACO is revealed through agents' actions, we formulate it as a sequential decision-making problem, naturally suited to multi-agent reinforcement learning (MARL) \citep{bernstein2002complexity, oliehoek2016concise, yu2022mappo}. Existing MARL approaches typically manage coordination via agent-centric representations \citep{sunehag2018vdn, guestrin2002coordinated, sukhbaatar2016commnet}. Alongside neural MACO solvers that resolve step-level conflicts during parallel decoding \citep{berto2025parco}, these methods broadly share a common limitation: they localize coordination at the agent or step level, without explicitly modeling the shared environmental structure that drives their interdependencies.

\begin{wrapfigure}{r}{0.7\linewidth}
    \centering
    \includegraphics[width=\linewidth]{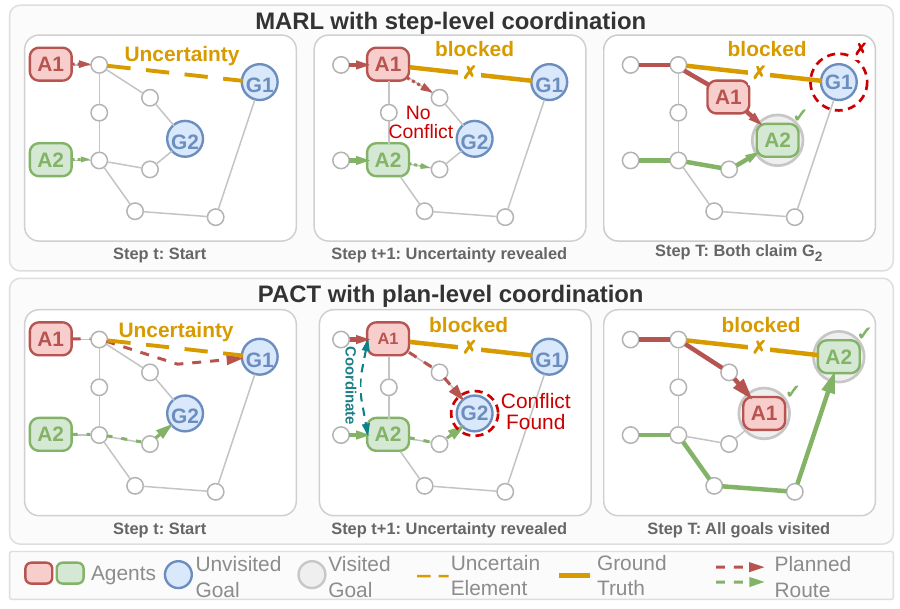}
    \caption{PACT with plan-level coordination prevents goal conflicts exhibited by MARL methods with step-level coordination.}
    \vspace{-9pt}
    \label{fig:motivation}
\end{wrapfigure}

In DMACO, coordination requirements are fundamentally induced by the environment. For instance, multiple agents might contend for a capacity-limited resource, or one agent's action might resolve local uncertainty crucial to another's future trajectory. Because these dependencies are grounded in the environment, relying solely on agent-level relations creates a \textit{structural mismatch}: the underlying graph elements driving the coordination are excluded from the representational mechanism. Furthermore, restricting coordination to the current step introduces a \textit{temporal mismatch}. An agent might commit to a trajectory that appears locally conflict-free, only to conflict with another agent's plan several steps later, as illustrated in Figure~\ref{fig:motivation}.

To address these structural and temporal mismatches, we introduce PACT (\textbf{P}lan-\textbf{A}ugmented \textbf{C}oordination via \textbf{T}race-graphs), a novel coordination framework for DMACO. First, each agent proposes multiple candidate plans. Every candidate plan is then attached to the environment graph via its projected \textit{trace}, a record of the graph elements on which it places demand. By performing coordination on this plan-augmented trace-graph, PACT enables the underlying environmental structure to explicitly mediate dependencies across agents' future candidate plans. We refer to this mechanism as \emph{environment-mediated coordination}. Our key contributions are as follows:
\begin{itemize}[leftmargin=*, nosep]
    \item We propose PACT, a novel coordination framework for DMACO problems that represents coordination requirements directly through the environmental structure that induces them, rather than through ungrounded relations defined between agents.
    \item We introduce the plan-augmented trace-graph, which attaches candidate plans to the elements they demand. The proposed message-passing scheme resolves contention over remaining capacities, thereby coordinating agents over future decision steps rather than only the current one.
    \item On three DMACO problems spanning routing and scheduling, we show that PACT outperforms neural MACO solvers and MARL methods in solution quality and zero-shot generalization on instances up to $64\times$ larger than the training set, at a fraction of the runtime of online replanning.
\end{itemize}

\section{Related Work}
\label{sec:related}

\noindent \textbf{Neural Multi-Agent Combinatorial Optimization.}
NCO amortizes the cost of solving recurring optimization problems by learning solution policies over instance distributions \citep{bengio2021machine, cappart2023combinatorial}. Constructive approaches generate solutions autoregressively \citep{kool2019attention, kwon2020pomo}, with multi-agent extensions coordinating agents' decisions to construct a joint solution. Sequential methods decode agents' actions in turn \citep{son2024equity, zheng2024dpn}, whereas parallel methods simultaneously decode agents' actions with inter-agent communication \citep{zong2022mapdp, hua2025camp}. These approaches focus on specific MACO problems, while PARCO \citep{berto2025parco} provides a general framework for parallel decoding across MACO problems, resolving conflicting selections through a learned priority-based handler. MACSIM \citep{luttmann2026macsim} learns through self-improvement to predict joint agent-action assignments that are conflict-free by construction. These methods assume that problem instances are fully specified before solution construction begins and coordinate agents at the current step. They therefore do not explicitly reason about information revealed during execution or dependencies between agents' decisions across future steps.

\noindent \textbf{Coordination in Multi-Agent Reinforcement Learning.}
Cooperative MARL typically employs centralized training with decentralized execution \citep{bernstein2002complexity, oliehoek2016concise}, under which representative algorithms such as multi-agent PPO (MAPPO) \citep{yu2022mappo} train a shared centralized critic, whereas independent PPO (IPPO) \citep{dewitt2020ippo} uses decentralized critics. Coordination is commonly achieved via value factorization, such as additive decomposition in VDN \citep{sunehag2018vdn} and monotonic mixing in QMIX \citep{rashid2018qmix}, explicit representations like deep coordination graphs (DCG) \citep{bohmer2020dcg}, or inter-agent communication \citep{sukhbaatar2016commnet, foerster2016dial, das2019tarmac, hu2024commformer}. In all these families, the coordination medium is agent-centric (e.g., agent utilities, pairwise payoffs, or directed messages). This introduces two limitations when these approaches are applied to DMACO. First, pairwise decompositions struggle to capture simultaneous multi-agent dependencies without incurring the exponential inference costs of higher-order factors \citep{guestrin2002coordinated, kok2006collaborative}. Second, these mechanisms coordinate the agents' actions at the current state, whereas conflicts in the future are reflected only implicitly through learned value estimates.

In DMACO settings, PACT addresses both limitations through environment-mediated coordination.
By connecting each candidate plan to the graph elements its trace occupies, PACT captures interactions between candidate plans, such as contention for shared capacity or dependence on unrevealed information, at the element that induces them.
This element-centric aggregation accommodates multi-agent dependencies without explicit agent-pair factors, while candidate traces expose coordination requirements spanning future decision steps.

\section{Preliminaries}
\label{sec:preliminaries}

We represent a DMACO problem as a graph, where each node is an \emph{element} that a solution selects, assigns, or sequences, and edges encode the relations among elements. We assume that any observation made by one agent is immediately available to the whole team, so that all agents act on the same information. The problem is therefore modeled as a fully cooperative multi-agent partially observable Markov decision process (MPOMDP) \citep{oliehoek2016concise}, defined by the tuple $\langle \mathcal{M}, \mathcal{S}, \{\mathcal{A}_i\}_{i \in \mathcal{M}}, \mathcal{T}, O, C, \gamma \rangle$, where $\mathcal{M}$ is the team of agents, $\mathcal{S}$ the states, $\{\mathcal{A}_i\}_{i \in \mathcal{M}}$ the action space for each agent, $\mathcal{T}$ the transition function, $O$ the observation function, $C$ the cost function, and $\gamma \in (0,1]$ the discount factor.
We factorize the state space $\mathcal{S}$ into observable graph states $\mathcal{G}_t$ and an unobserved latent realization $\xi$, with $O(\mathcal{G}_t, \xi) = \mathcal{G}_t$.
At the start of each episode we sample $\xi \sim p(\xi)$, and it is held fixed for the episode.
The observed graph $\mathcal{G}_t = (\mathcal{V}_t, \mathcal{E}_t, X_t)$ consists of the elements $\mathcal{V}_t$, the relations $\mathcal{E}_t$ among them, and the attributes $X_t$ of each element and each agent. The attributes $X_t$ also determine the set $\mathcal{M}_t \subseteq \mathcal{M}$ of agents active at decision step $t$. 
Each active agent $i \in \mathcal{M}_t$ selects an action $a_{i,t} \in \mathcal{A}_i(\mathcal{G}_t)$ from its own policy $\pi_i$ conditioned on the graph observation $\mathcal{G}_t$, where $\mathcal{A}_i(\mathcal{G}_t) \subseteq \mathcal{V}_t \cup \{\varnothing\}$, so that agent $i$ commits either to a single element or to none. We write $\pi = \{\pi_i\}_{i \in \mathcal{M}}$ for the team policy. The state then advances according to $\mathcal{G}_{t+1} \sim \mathcal{T}(\cdot \mid \mathcal{G}_t, \mathbf{a}_t, \xi)$, where the joint action $\mathbf{a}_t = \{a_{i,t}\}_{i \in \mathcal{M}_t}$. 
Each decision step incurs a cost $C(\mathcal{G}_t, \mathbf{a}_t, \xi) \in \mathbb{R}_{\geq0}$, and the agents minimize the expected discounted cost $J(\pi) = \mathbb{E}_{\xi, \pi} \Big[ \textstyle\sum_{t} \gamma^t \, C(\mathcal{G}_t, \mathbf{a}_t, \xi) \Big]$.

Capacities and demands serve as the general interface for defining resource constraints in combinatorial optimization \citep{blazewicz1983scheduling} and constraint programming \citep{aggoun1993extending}.
We assign a capacity $\kappa_t(v) \ge 0$ to each element $v \in \mathcal{V}_t$, recorded in the attributes ${X}_t$.
The action $a_i$ from agent $i$ places a demand $d(v, a_{i,t}) \ge 0$ on the element $v$.
A unit capacity $\kappa(v) = 1$ makes an element exclusive, which covers a goal location claimed by a single traveler \citep{toth2014vehicle} and a job processed by a single machine \citep{pinedo2012scheduling}, while larger capacities describe resources that several agents may hold simultaneously. 
At decision step $t$, the demand on element $v$ from all active agents is $d_t(v) = \sum_{i \in \mathcal{M}_t} d(v, a_{i,t})$. 
The joint actions permitted by the remaining capacities form the set
\begin{equation}
    \mathcal{A}(\mathcal{G}_t) = \left\{ a = \{a_i\}_{i \in \mathcal{M}_t} \in
\prod_{i \in \mathcal{M}_t} \mathcal{A}_i(\mathcal{G}_t) \ : \ d_t(v) \le \kappa_t(v)
\text{ for every } v \in \mathcal{V}_t \right\}.
    \label{eq:capacity}
\end{equation}
Eq.~\eqref{eq:capacity} is
the only feasibility constraint that couples the agents, and it is a conjunction of conditions taken
element by element.

\section{Methodology}
\label{sec:method}

\begin{figure}[t]
\centering
\includegraphics[width=\linewidth]{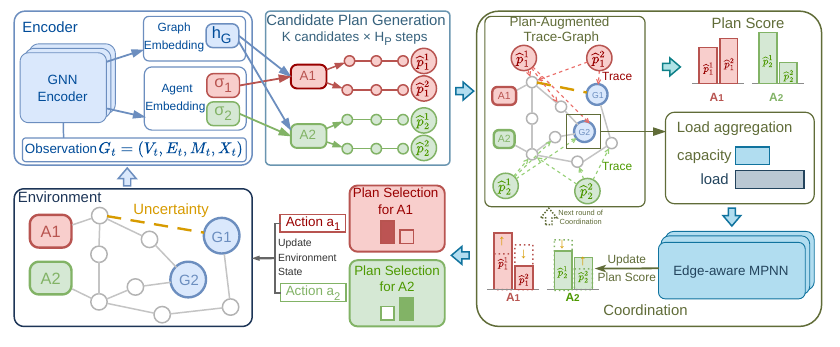}
\caption{Overview of PACT. After encoding graph observations, agents independently generate candidate plans to form a plan-augmented trace-graph for multi-step coordination. The joint action $\mathbf{a}_t$ comprises the initial step of each selected plan.
}
\vspace{-9pt}
\label{fig:overview}
\end{figure}

We now present PACT, whose overall architecture is illustrated in Figure~\ref{fig:overview}.
Each agent first generates a set of candidate plans instead of committing to a
single next action (\Cref{sec:candidates}). The candidate plans are incorporated into the graph
through the elements they occupy, forming a plan-augmented trace-graph over which inter-agent
coordination is performed (\Cref{sec:graph}). Each agent subsequently selects one of its
coordinated candidate plans and executes its first action (\Cref{sec:plan-selection}). Finally, all modules are trained end to end (\Cref{sec:training}). The complete
forward pass of PACT is summarized in Algorithm~\ref{alg:pact}.

\subsection{Candidate Plan Generation}
\label{sec:candidates}
As was illustrated in Figure~\ref{fig:motivation}, coordination over individual actions is inherently myopic: agents whose current actions are compatible may still contend for the same resources several steps ahead. PACT therefore coordinates over short-horizon candidate plans. However, planning in the joint action space scales exponentially with both the number of agents and the planning horizon \citep{bernstein2002complexity,kearns2002sparse}. We instead decouple plan proposal from multi-agent coordination: each agent independently generates a compact set of candidate plans, while interactions among these plans are handled by the coordination mechanism in \Cref{sec:graph}.

Given the current graph observation $\mathcal{G}_t$, a GNN-based encoder $\mathrm{Enc}_\theta$, whose parameters $\theta$ are shared with every other learned module of PACT, extracts graph-related representations $h_G$ together with an agent-related representation $\sigma_i$ for each active agent,
\begin{equation}
    \big(
        h_G,\;
        \{\sigma_i\}_{i \in \mathcal{M}_t}
    \big)
    =
    \mathrm{Enc}_\theta(\mathcal{G}_t).
    \label{eq:encode}
\end{equation}
The graph representation $h_G$ describes the decision-relevant structure and elements of the current problem instance, while $\sigma_i$ provides the agent-specific context needed to specialize the graph representation to agent $i$. 

For each agent $i$, PACT generates $K$ candidate plans, each consisting of a sequence of $H_p$ actions,
\begin{equation}
    \hat{p}_i^k
    =
    \big(
        \hat{a}_{i,t}^k,\;
        \hat{a}_{i,t+1}^k,\;
        \ldots,\;
        \hat{a}_{i,t+H_p-1}^k
    \big),
    \qquad
    k=1,\ldots,K.
    \label{eq:candidate}
\end{equation}

Each candidate plan is constructed through a rollout governed by a learned candidate plan generation policy $\pi_\theta^{\mathrm{plan}}$. 
To ensure diversity among the $K$ candidate plans, their first actions are drawn jointly without replacement from the policy at the initial rollout state $\pi_\theta^{\mathrm{plan}}(\mathcal{G}_t)$.
Following this initialization, each of the $K$ plans unrolls independently.
At each rollout step $j$, $\pi_\theta^{\mathrm{plan}}$ conditions on the encoded graph $h_G$, the agent representation $\sigma_i$, and the rollout state at step $j$, $\hat{\mathcal G}_{t+j,i}^k$ to extend the partial plan, ensuring that subsequent decisions dynamically reflect the accumulated consequences of preceding actions.

\subsection{Plan-Augmented Trace-Graph and Coordination}
\label{sec:graph}

Recall the routing scenario shown in Figure~\ref{fig:motivation}. In this scenario, two agents move apart at the current step but then both claim
the same goal several steps later. Their current actions place demand on different elements, so
Eq.~\eqref{eq:capacity} holds at every step until they arrive, and the demand that must be
reconciled is located at the goal, where the final two actions together exceed its remaining capacity $\kappa_t(v)$. PACT
therefore records, for each candidate plan, which elements it would occupy, the induced demand, and the step of the plan. We refer to this record as the candidate plan's
\emph{trace},
\begin{equation}
    \mathrm{Tr}(\hat{p}_i^k) \;=\; \big\{ \, \big(v, \; d(v, \hat{a}_{i,t+j}^k), \;
    j \big) \;:\; d(v, \hat{a}_{i,t+j}^k) > 0, \;\;
    j = 0, \ldots, H_p-1 \, \big\} ,
    \label{eq:trace}
\end{equation}
where $j$ indexes the step of the plan at which the demand is placed.

Once $K$ candidate plans are generated in parallel, their traces form the \emph{plan-augmented trace-graph} $\chi_t \;=\; \big( \mathcal{V}_t \cup \hat{\mathcal{V}}_t, \;\;
    \mathcal{E}_t \cup \hat{\mathcal{E}}_t \big)$, where candidate plan nodes $\hat{\mathcal{V}}_t = \{ \hat{p}_i^k : i \in \mathcal{M}_t, \,
k \le K \}$ and trace arcs $\hat{\mathcal{E}}_t = \{ (\hat{p}, v) : (v, \cdot, \cdot) \in
\mathrm{Tr}(\hat{p}) \}$. Each arc carries the demand and the plan step that
produced it. Arcs join candidate plans to elements but not to one another. This forces information exchange between agents to occur through the graph structure. Thus, the same graph structure captures both the original combinatorial problem and the prospective interactions introduced by the plans. 

PACT coordinates the candidate plans by iteratively alternating between two operations in order to compute a set of \textit{preferences}. First, the resource contention induced by the agents' current plan preferences is estimated. Second, preferences are updated through message passing on the plan-augmented trace-graph $\chi_t$. Concretely, let
$s_{i}^{(r)}(k)$ denote the \textit{score} of plan $k$ of agent $i$ after coordination round $r=0,\ldots,R$. The corresponding \textit{preference} is:
\begin{equation}
    q_{i}^{(r)}(k)
    =
    \frac{
        \exp(s_{i}^{(r)}(k))
    }{
        \sum_{k'=1}^{K}\exp(s_{i}^{(r)}(k'))
    }.
    \label{eq:plan_preference}
\end{equation}
Every node of $\chi_t$ carries a representation $h^{(r)}$ at round $r$, whose initial value $h^{(0)}$ is given by $h_G$ for element nodes and by an encoding of the node's trace for candidate plan nodes. A scoring head $f_{\mathrm{score}}$ turns a candidate plan's representation into its score, so that $s_{i}^{(0)}(k) = \eta\,\tanh\!\big(f_{\mathrm{score}}(h^{(0)}(\hat{p}_i^k))\big)$ before information is exchanged between agents, where $\eta>0$ is a fixed step size bound.

Given the current preferences $q_{i}^{(r)}$, each element aggregates the demand induced by all candidate plans that reach it. We define $d(v,\hat{p}_i^k) = \sum_{j} d(v, \hat{a}_{i,t+j}^k)$ as the total demand that $\hat{p}_i^k$ places on $v$ across its steps. The
load on element $v$ at round $r$ is the preference-weighted demand of all plans,
\begin{equation}
    L^{(r)}(v)
    =
    \sum_{i\in\mathcal{M}_t}
    \sum_{k=1}^{K}
    q_{i}^{(r)}(k)\,
    \times d(v,\hat{p}_i^k).
    \label{eq:load}
\end{equation}
Hence, $L^{(r)}(v)$ measures the expected demand induced by the agents' current preferences $q_{i}^{(r)}$ over their candidate plans. 
The load is compared with the remaining capacity $\kappa_t(v)-d_t(v)$ of each element and injected into its representation through a learned projection $W_{\mathrm{load}}$, normalized by the capacity $\kappa_t(v)$,
\begin{equation}
    \tilde{h}^{(r)}(v)
    =
    h^{(r)}(v)
    +
    W_{\mathrm{load}}
    \left[
        \frac{L^{(r)}(v)}{\kappa_t(v)},\;
        \frac{
            \mathrm{ReLU}\!\left(L^{(r)}(v)-(\kappa_t(v)-d_t(v))\right)
        }{
            \kappa_t(v)
        }
    \right],
    \label{eq:inject}
\end{equation}
where $\frac{L^{(r)}(v)}{\kappa_t(v)}$ describes resource utilization, while $\left(L^{(r)}(v)-(\kappa_t(v) - d_t(v))\right)$ explicitly identifies demand beyond the remaining capacity. Normalizing by capacity provides a common scale for resources with different capacities.

An edge-aware message-passing layer, shared across rounds, then propagates this information over the plan-augmented trace-graph,

\begin{align}
    h^{(r+1)}
    &=
    \mathrm{MPNN}
    \big(
        \tilde{h}^{(r)},\chi_t
    \big),
    \label{eq:mp}
    \\
    s_{i}^{(r+1)}(k)
    &=
    s_{i}^{(r)}(k)
    +
    \eta\,\tanh\!\big(
        f_{\mathrm{score}}(h^{(r+1)}(\hat{p}_i^k))
    \big).
    \label{eq:score}
\end{align}

The updated scores are renormalized over the $K$ candidate plans of each agent to obtain $q_{i}^{(r+1)}$, which in turn induces a new load profile through Eq.~\eqref{eq:load}. The candidate plan scores are therefore updated iteratively through coordination,
allowing an agent's preference over its own plans to respond to the evolving preferences of the rest of the team.

\subsection{Plan Selection and Execution}
\label{sec:plan-selection}

After $R$ coordination rounds, each agent selects a candidate plan based on its final coordinated scores, $z_i \sim \operatorname{softmax}( s_{i}^{(R)}/\tau )$,
where $\tau$ is a learned temperature controlling the sharpness of selection over candidate plans. Each agent then executes the first action of its chosen plan, ${a_{i,t} = \hat{p}_i^{z_i}[0]}$.
The resulting joint action $\mathbf{a}_t = \{a_{i,t}\}_{i\in\mathcal{M}_t}$ is applied to the environment via $\mathcal{T}$, which incurs the cost $C(\mathcal{G}_t, \mathbf{a}_t, \xi)$, reveals the components of $\xi$ associated with $\mathbf{a}_t$, and yields the subsequent graph observation $\mathcal{G}_{t+1}$. Following the receding-horizon principle, PACT regenerates and coordinates new candidate plans from $\mathcal{G}_{t+1}$ at each step. Consequently, while future actions within a candidate plan serve as look-ahead information for multi-agent coordination, execution remains closed-loop and adapts to newly revealed information.

\begin{algorithm}[t]
\caption{PACT forward pass at decision step $t$}
\label{alg:pact}
\begin{algorithmic}[1]
\STATE \textbf{Input:} graph observation $\mathcal{G}_t$
\STATE Encode $\mathcal{G}_t$ once into $h_G$ and $\{\sigma_i\}$ \hfill (Eq.~\eqref{eq:encode})
\FOR{each agent $i \in \mathcal{M}_t$ (in parallel)}
    \STATE Generate $K$ candidate plans $\hat{p}_i^1,\ldots,\hat{p}_i^K$ through rollout, with first actions drawn jointly without replacement (\Cref{sec:candidates})
\ENDFOR
\STATE Construct the plan-augmented trace-graph $\chi_t$ from the candidate plan traces \hfill (Eq.~\eqref{eq:trace})
\STATE Initialize candidate plan scores $s_{i}^{(0)}$ and preferences $q_{i}^{(0)}=\mathrm{softmax}(s_{i}^{(0)})$ \hfill (Eq.~\eqref{eq:plan_preference})
\FOR{coordination round $r = 0,\ldots,R-1$}
    \STATE Compute the load $L^{(r)}(v)$ induced by the current plan preferences \hfill (Eq.~\eqref{eq:load})
    \STATE Inject capacity-aware load features and perform message passing over $\chi_t$ \hfill (Eqs.~\eqref{eq:inject} and~\eqref{eq:mp})
    \STATE Update candidate plan scores $s_{i}^{(r+1)}$ and preferences $q_{i}^{(r+1)}$ (\Cref{sec:graph}) \hfill (Eq.~\eqref{eq:score})
\ENDFOR
\FOR{each agent $i \in \mathcal{M}_t$ (in parallel)}
    \STATE Sample $z_i \sim \mathrm{softmax}(s_{i}^{(R)}/\tau)$ and set $a_{i,t} = \hat{p}_i^{z_i}[0]$ (\Cref{sec:plan-selection})
\ENDFOR
\STATE \textbf{Output:} Joint action $\mathbf{a}_t = \{a_{i,t}\}_{i \in \mathcal{M}_t}$
\end{algorithmic}
\end{algorithm}

\subsection{Policy Training}
\label{sec:training}
PACT learns its plan generation and plan-selection policies from the return. Unlike a conventional policy in which the sampled policy output is the environment action, PACT produces each agent's action through a structured sequence of stochastic decisions: candidate plans are first generated and subsequently selected after coordination. The policy objective therefore needs to account for the internal decisions that give rise to the executed action.

We decompose the policy update into three components: \emph{candidate plan proposal} ($\mathrm{prop}$), \emph{plan rollout} ($\mathrm{roll}$), and \emph{plan selection} ($\mathrm{sel}$). For each component $g \in \{\mathrm{prop},\mathrm{roll},\mathrm{sel}\}$, we optimize the clipped surrogate objective:
\begin{equation}
    \mathcal{L}_{g}(\theta)
    =
    \mathbb{E}_t
    \left[
        \sum_{i \in \mathcal{M}_t}
        \min
        \left(
            \rho_{i,t}^{g}(\theta)\hat{A}_t,\,
            \operatorname{clip}
            \big(
                \rho_{i,t}^{g}(\theta),
                1-\epsilon_{g},
                1+\epsilon_{g}
            \big)\hat{A}_t
        \right)
    \right],
    \label{eq:component_ppo}
\end{equation}
where $\rho_{i,t}^{g}$ is the importance ratio between the probabilities that the current and the behavior policies assign to agent $i$'s decisions in component $g$. For $\mathrm{prop}$, these decisions are the $K$ first actions, and the ratio compares their joint probability as an ordered set drawn without replacement. For $\mathrm{roll}$, they are the actions at steps $j \ge 1$ of the selected plan $\hat{p}_i^{z_i}$. For $\mathrm{sel}$, the decision is the selected index $z_i$. The advantage $\hat{A}_t$ is estimated with generalized advantage estimation \citep{schulman2015high} from a critic conditioned on the graph observation $\mathcal{G}_t$, and $\epsilon_{g}$ is the clipping width. The overall objective sums the component objectives as
$    \mathcal{L}_{\mathrm{PACT}}(\theta)
    =
    \mathcal{L}_{\mathrm{prop}}(\theta)
    +
    \mathcal{L}_{\mathrm{roll}}(\theta)
    +
    \mathcal{L}_{\mathrm{sel}}(\theta)$.
Clipping each component separately preserves PACT's structured hierarchy while allowing a single return signal to jointly train proposal, rollout, and selection. 

\begingroup
\setlength{\textfloatsep}{5pt}
\setlength{\intextsep}{5pt}
\section{Experiments}
\label{sec:experiments}

\subsection{Experimental Setting}
\label{sec:setting}

\noindent \textbf{Problems.}
We evaluate PACT on three representative DMACO problems, spanning both routing and scheduling domains.
In the \textbf{Multi-Agent Canadian Traveller Problem (MACTP)} \citep{papadimitriou1991shortest, shiri2019competitive}, a team traverses a graph to collectively visit goal nodes at minimum total travel cost, with edge traversability revealed upon reaching an endpoint or attempting a crossing.
Travelers are agents and goal nodes have unit capacity.
In the \textbf{Dynamic Flexible Job-Shop Problem (DFJSP)} \citep{ouelhadj2009survey}, jobs comprise precedence-constrained operations assigned to eligible machines; whether an uncertain machine is operational or broken is revealed when an operation is dispatched to it.
In the \textbf{Non-Clairvoyant Flexible Flow Shop Problem (NCFFSP)} \citep{motwani1994nonclairvoyant}, jobs pass through ordered stages of interchangeable machines whose latent processing-time factors are revealed upon dispatch.
Both scheduling problems minimize makespan, with machines acting as agents and jobs as unit-capacity elements.
Appendix~\ref{app:problem-defs} provides the formal models and revelation rules.

\noindent \textbf{Baselines.}
We compare PACT with two multi-agent NCO methods, PARCO and MACSIM, which are evaluated as centralized joint-construction solvers through task wrappers, retaining their original training objectives and model configurations. For PACT, PARCO, and MACSIM two decoding strategies named \textit{greedy (g.)} and \textit{sampling (s.)} are used. We also compare four cooperative MARL baselines (MAPPO, IPPO, VDN, and DCG), using matched graph encoders over the common observation interface and shared parameters across agents.
MAPPO, IPPO, and VDN produce per-agent actions, whereas DCG coordinates joint actions through max-plus message passing on an agent coordination graph. PACT uses a critic conditioned on the graph observation and performs decentralized execution with communication through the plan-augmented trace-graph, in which each agent selects its own candidate plan.
Gurobi \citep{gurobi} and OR-Tools \citep{ortools} provide online replanning baselines and oracle references.
Appendix~\ref{app:baseline-methods} details the baseline implementations and adaptations.

\noindent \textbf{Training.}
For each problem, all learning-based methods are trained on the same instance set with a mixture of small sizes.
The mixture spans $n\in\{10,13,16\}$ nodes for MACTP, $N\in\{12,16,20\}$ jobs for DFJSP, and $N\in\{12,16,20\}$ jobs with $3$ stages and $3$ or $4$ machines per stage for NCFFSP.
Dataset generation seeds are fixed, and each method is trained with $10$ independent seeds.
The training configurations are given in Appendix~\ref{app:baselines}.

\noindent \textbf{Evaluation.}
All methods use the same $100$ test instances per configuration, and learning-based results are averaged across the $10$ training seeds.
Instance sizes are denoted by nodes $\times$ agents $\times$ goals ($n\times A\times G$) for MACTP, jobs $\times$ machines $\times$ operations ($N\times M\times O$) for DFJSP, and jobs $\times$ machines per stage $\times$ stages ($N\times M_\ell\times S$) for NCFFSP.
The first configuration in each problem block is in-distribution (ID), matching the training mixture's largest size; the remaining four evaluate zero-shot transfer (ZST) to larger instances.
The final two NCFFSP configurations also increase the stage count from $3$ to $4$.
Appendix~\ref{app:complete-results} details the evaluation protocols.

\begin{table*}[t]
\caption{Results on three DMACO problems under ID and ZST evaluation. Cost $\downarrow$, Time (s) $\downarrow$, and Success (\%) $\uparrow$. N/A marks unsupported configurations, T/O marks runs without a feasible solution within the time limit, and OOM marks runs that exceed GPU memory.}
\label{tab:main}
\setlength{\aboverulesep}{0pt}
\setlength{\belowrulesep}{0pt}
\setlength{\tabcolsep}{2pt}
\renewcommand{\arraystretch}{1.12}
\begin{adjustbox}{width=0.98\textwidth, center}
\begin{tabular}{c|l| c c c | c c c | c c c | c c c | c c c}
\toprule[0.6mm]
\multicolumn{17}{c}{\textbf{MACTP}} \\
\toprule
\multicolumn{2}{c|}{$n \times A \times G$} & \multicolumn{3}{c|}{16$\times$4$\times$6 (ID)} & \multicolumn{3}{c|}{32$\times$6$\times$10 (ZST)} & \multicolumn{3}{c|}{128$\times$12$\times$24 (ZST)} & \multicolumn{3}{c|}{512$\times$24$\times$48 (ZST)} & \multicolumn{3}{c}{1024$\times$32$\times$64 (ZST)} \\
\midrule
\multicolumn{2}{c|}{Metric} & Cost & Succ. & Time & Cost & Succ. & Time & Cost & Succ. & Time & Cost & Succ. & Time & Cost & Succ. & Time \\
\midrule
\multirow{2}{*}{Oracle} & \textit{Gurobi (10m)} & \textit{2.84} & \textit{100.0} & \textit{0.18} & \textit{4.28} & \textit{100.0} & \textit{0.25} & \textit{9.52} & \textit{100.0} & \textit{386.5} & \textit{21.9} & \textit{100.0} & \textit{627.2} & \textit{3132.5} & \textit{93.0} & \textit{652.3} \\
 & \textit{OR-Tools (10m)} & \textit{2.84} & \textit{100.0} & \textit{0.66} & \textit{4.28} & \textit{100.0} & \textit{37.5} & \textit{13.0} & \textit{100.0} & \textit{624.7} & \textit{917.7} & \textit{97.0} & \textit{633.8} & \textit{T/O} & \textit{T/O} & \textit{T/O} \\
\midrule
\multirow{2}{*}{Replan} & Gurobi       & 5.02 & 90.0 & 0.42 & 10.1 & 79.0 & 1.06 & 95.0 & 37.0 & 13.9 & T/O & T/O & T/O & T/O & T/O & T/O \\
 & OR-Tools     & 4.08 & 93.0 & 0.92 & 23.5 & 34.0 & 6.37 & 231.9 & 39.0 & 121.3 & T/O & T/O & T/O & T/O & T/O & T/O \\
\midrule
\multirow{7}{*}{Greedy} & MAPPO (\textit{g.}) & 3.83 & 96.8 & 0.027 & 6.78 & 98.1 & 0.020 & 32.8 & 95.7 & 0.79 & OOM & OOM & OOM & OOM & OOM & OOM \\
 & IPPO (\textit{g.}) & 3.92 & 97.1 & 0.015 & 6.84 & 98.8 & 0.033 & 28.6 & 97.2 & 0.63 & OOM & OOM & OOM & OOM & OOM & OOM \\
 & VDN (\textit{g.}) & \textbf{3.52} & 98.9 & 0.012 & \underline{5.84} & 99.6 & 0.019 & \underline{20.7} & 99.9 & 0.38 & OOM & OOM & OOM & OOM & OOM & OOM \\
 & DCG (\textit{g.}) & 3.68 & 99.3 & 0.011 & 8.44 & 98.1 & 0.048 & 70.0 & 93.3 & 0.99 & OOM & OOM & OOM & OOM & OOM & OOM \\
 & PARCO (\textit{g.}) & 4.56 & 98.9 & 0.004 & 10.8 & 95.0 & 0.017 & 195.1 & 41.2 & 0.11 & 1722.0 & 4.2 & 0.35 & 3999.2 & 4.5 & 0.73 \\
 & MACSIM (\textit{g.}) & 7.36 & 89.9 & 0.038 & 12.7 & 95.2 & 0.035 & 34.8 & 99.6 & 0.047 & \underline{90.2} & 100.0 & 0.18 & \underline{131.3} & 100.0 & 2.20 \\
 & PACT (\textit{g.}) & \underline{3.66} & 99.1 & 0.056 & \textbf{5.79} & 99.6 & 0.064 & \textbf{18.6} & 99.7 & 0.34 & \textbf{72.3} & 99.0 & 2.61 & \textbf{86.1} & 99.6 & 11.0 \\
\midrule
\multirow{3}{*}{Sampling} & PARCO (\textit{s.}) & 3.35 & 100.0 & 0.094 & \underline{5.50} & 100.0 & 0.15 & \underline{17.8} & 99.7 & 2.67 & 372.6 & 86.3 & 9.83 & 1341.6 & 71.3 & 27.5 \\
 & MACSIM (\textit{s.}) & \underline{3.27} & 100.0 & 0.48 & 6.32 & 100.0 & 0.68 & 24.0 & 100.0 & 1.91 & \underline{84.2} & 100.0 & 42.0 & \underline{125.8} & 100.0 & 381.7 \\
 & PACT (\textit{s.}) & \textbf{2.89} & 100.0 & 5.27 & \textbf{4.66} & 100.0 & 5.45 & \textbf{15.5} & 100.0 & 10.5 & \textbf{50.3} & 100.0 & 51.9 & \textbf{84.0} & 100.0 & 136.9 \\
\bottomrule[0.5mm]
\addlinespace[1mm]
\multicolumn{17}{c}{\textbf{DFJSP}} \\
\toprule
\multicolumn{2}{c|}{$N \times M \times O$} & \multicolumn{3}{c|}{20$\times$8$\times$6 (ID)} & \multicolumn{3}{c|}{30$\times$14$\times$6 (ZST)} & \multicolumn{3}{c|}{50$\times$16$\times$8 (ZST)} & \multicolumn{3}{c|}{80$\times$20$\times$8 (ZST)} & \multicolumn{3}{c}{120$\times$24$\times$10 (ZST)} \\
\midrule
\multicolumn{2}{c|}{Metric} & Cost & Succ. & Time & Cost & Succ. & Time & Cost & Succ. & Time & Cost & Succ. & Time & Cost & Succ. & Time \\
\midrule
\multirow{2}{*}{Oracle} & \textit{Gurobi (10m)} & \textit{154.9} & \textit{100.0} & \textit{624.6} & \textit{133.7} & \textit{100.0} & \textit{626.3} & \textit{291.7} & \textit{100.0} & \textit{614.2} & \textit{T/O} & \textit{T/O} & \textit{T/O} & \textit{T/O} & \textit{T/O} & \textit{T/O} \\
 & \textit{OR-Tools (10m)} & \textit{148.3} & \textit{100.0} & \textit{624.2} & \textit{126.1} & \textit{100.0} & \textit{624.2} & \textit{221.6} & \textit{100.0} & \textit{600.6} & \textit{T/O} & \textit{T/O} & \textit{T/O} & \textit{T/O} & \textit{T/O} & \textit{T/O} \\
\midrule
\multirow{2}{*}{Replan} & Gurobi       & \textbf{156.8} & 100.0 & 28.8 & T/O & T/O & T/O & T/O & T/O & T/O & T/O & T/O & T/O & T/O & T/O & T/O \\
 & OR-Tools     & 172.7 & 87.0 & 61.6 & 149.4 & 100.0 & 76.3 & T/O & T/O & T/O & T/O & T/O & T/O & T/O & T/O & T/O \\
\midrule
\multirow{7}{*}{Greedy} & MAPPO (\textit{g.}) & 170.2 & 100.0 & 0.022 & 148.4 & 100.0 & 0.021 & \underline{245.1} & 100.0 & 0.070 & \underline{299.0} & 100.0 & 0.24 & \underline{429.2} & 100.0 & 0.84 \\
 & IPPO (\textit{g.}) & 207.7 & 96.1 & 0.051 & 156.1 & 99.7 & 0.054 & 295.6 & 98.7 & 0.24 & 380.1 & 98.6 & 0.87 & 584.4 & 98.5 & 3.45 \\
 & VDN (\textit{g.}) & 632.1 & 54.4 & 0.056 & 688.7 & 66.5 & 0.076 & 1641.3 & 57.7 & 0.33 & 2498.4 & 59.6 & 1.27 & 4311.0 & 59.9 & 5.54 \\
 & DCG (\textit{g.}) & 986.6 & 15.1 & 0.088 & 1476.8 & 13.4 & 0.13 & 3089.9 & 7.4 & 0.61 & 5017.7 & 5.4 & 2.72 & 9296.8 & 0.5 & 13.1 \\
 & PARCO (\textit{g.}) & 170.5 & 100.0 & 0.004 & \underline{146.0} & 100.0 & 0.004 & 258.4 & 99.7 & 0.015 & 336.2 & 97.4 & 0.054 & 2578.1 & 7.6 & 0.35 \\
 & MACSIM (\textit{g.}) & 608.3 & 54.7 & 0.18 & 596.9 & 70.1 & 0.21 & 1241.3 & 60.2 & 0.48 & 1305.2 & 58.0 & 0.84 & 1866.3 & 61.2 & 1.64 \\
 & PACT (\textit{g.}) & \underline{162.5} & 100.0 & 0.093 & \textbf{141.5} & 100.0 & 0.10 & \textbf{233.2} & 100.0 & 0.25 & \textbf{283.9} & 100.0 & 0.77 & \textbf{398.4} & 100.0 & 2.41 \\
\midrule
\multirow{3}{*}{Sampling} & PARCO (\textit{s.}) & 165.0 & 100.0 & 0.087 & 142.8 & 100.0 & 0.096 & 249.7 & 100.0 & 0.18 & 314.5 & 100.0 & 0.28 & 490.3 & 100.0 & 2.30 \\
 & MACSIM (\textit{s.}) & \underline{156.2} & 100.0 & 0.20 & \underline{136.4} & 100.0 & 0.25 & \underline{242.3} & 100.0 & 0.93 & \underline{307.2} & 100.0 & 1.83 & \underline{458.3} & 100.0 & 5.24 \\
 & PACT (\textit{s.}) & \textbf{155.9} & 100.0 & 10.5 & \textbf{134.1} & 100.0 & 10.5 & \textbf{232.4} & 100.0 & 23.6 & \textbf{279.6} & 100.0 & 54.4 & \textbf{396.5} & 100.0 & 112.1 \\
\bottomrule[0.5mm]
\addlinespace[1mm]
\multicolumn{17}{c}{\textbf{NCFFSP}} \\
\toprule
\multicolumn{2}{c|}{$N \times M_\ell \times S$} & \multicolumn{3}{c|}{20$\times$4$\times$3 (ID)} & \multicolumn{3}{c|}{30$\times$6$\times$3 (ZST)} & \multicolumn{3}{c|}{50$\times$7$\times$3 (ZST)} & \multicolumn{3}{c|}{80$\times$6$\times$4 (ZST)} & \multicolumn{3}{c}{120$\times$8$\times$4 (ZST)} \\
\midrule
\multicolumn{2}{c|}{Metric} & Cost & Succ. & Time & Cost & Succ. & Time & Cost & Succ. & Time & Cost & Succ. & Time & Cost & Succ. & Time \\
\midrule
\multirow{2}{*}{Oracle} & \textit{Gurobi (10m)} & \textit{34.3} & \textit{100.0} & \textit{624.5} & \textit{30.7} & \textit{100.0} & \textit{600.2} & \textit{41.2} & \textit{100.0} & \textit{601.6} & \textit{91.2} & \textit{100.0} & \textit{602.2} & \textit{183.5} & \textit{100.0} & \textit{607.3} \\
 & \textit{OR-Tools (10m)} & \textit{32.0} & \textit{100.0} & \textit{124.0} & \textit{26.9} & \textit{100.0} & \textit{603.6} & \textit{36.0} & \textit{100.0} & \textit{600.2} & \textit{67.5} & \textit{100.0} & \textit{600.3} & \textit{70.8} & \textit{100.0} & \textit{600.2} \\
\midrule
\multirow{2}{*}{Replan} & Gurobi       & \underline{37.5} & 100.0 & 15.7 & \textbf{34.1} & 100.0 & 19.3 & 225.1 & 55.0 & 31.4 & T/O & T/O & T/O & T/O & T/O & T/O \\
 & OR-Tools     & \textbf{37.1} & 100.0 & 6.62 & 36.5 & 100.0 & 13.6 & 53.6 & 100.0 & 32.2 & 351.3 & 96.0 & 112.5 & T/O & T/O & T/O \\
\midrule
\multirow{7}{*}{Greedy} & MAPPO (\textit{g.}) & 46.5 & 97.4 & 0.019 & 38.0 & 99.2 & 0.020 & \underline{49.4} & 99.3 & 0.065 & \underline{94.5} & 98.2 & 0.33 & \underline{93.8} & 98.4 & 1.13 \\
 & IPPO (\textit{g.}) & 47.3 & 98.5 & 0.017 & 39.3 & 99.3 & 0.020 & 52.6 & 98.9 & 0.070 & 99.8 & 97.8 & 0.30 & 97.1 & 98.6 & 1.05 \\
 & VDN (\textit{g.}) & 50.0 & 97.8 & 0.017 & 44.2 & 96.8 & 0.027 & 63.9 & 93.2 & 0.12 & 151.0 & 84.4 & 1.04 & 186.1 & 81.6 & 4.62 \\
 & DCG (\textit{g.}) & 84.1 & 71.7 & 0.036 & 102.4 & 58.9 & 0.069 & 152.4 & 59.0 & 0.26 & 376.7 & 50.3 & 1.64 & 513.9 & 48.4 & 6.98 \\
 & PARCO (\textit{g.}) & 42.9 & 99.3 & 0.005 & 49.8 & 98.8 & 0.009 & 66.8 & 98.8 & 0.010 & N/A & N/A & N/A & N/A & N/A & N/A \\
 & MACSIM (\textit{g.}) & 64.7 & 82.9 & 0.067 & 55.4 & 91.7 & 0.057 & 98.4 & 86.6 & 0.11 & 212.6 & 82.7 & 0.29 & 213.9 & 89.1 & 0.35 \\
 & PACT (\textit{g.}) & 41.5 & 100.0 & 0.049 & \underline{35.9} & 100.0 & 0.063 & \textbf{48.8} & 99.9 & 0.17 & \textbf{83.4} & 99.7 & 0.51 & \textbf{86.7} & 99.4 & 1.84 \\
\midrule
\multirow{3}{*}{Sampling} & PARCO (\textit{s.}) & \underline{37.3} & 100.0 & 0.085 & 39.6 & 100.0 & 0.099 & 57.4 & 100.0 & 0.13 & N/A & N/A & N/A & N/A & N/A & N/A \\
 & MACSIM (\textit{s.}) & 39.0 & 100.0 & 0.21 & \underline{36.1} & 100.0 & 0.32 & \underline{50.7} & 100.0 & 0.71 & \underline{97.3} & 100.0 & 2.09 & \underline{109.6} & 100.0 & 3.87 \\
 & PACT (\textit{s.}) & \textbf{36.2} & 100.0 & 5.35 & \textbf{35.4} & 100.0 & 6.17 & \textbf{44.8} & 100.0 & 11.7 & \textbf{82.5} & 100.0 & 32.5 & \textbf{85.6} & 100.0 & 97.0 \\
\bottomrule[0.5mm]
\end{tabular}
\end{adjustbox}
\end{table*}

\subsection{Main Results}
\label{sec:main-results}

Table~\ref{tab:main} shows that PACT, using the same framework across all three problems with decentralized execution and coordination through the plan-augmented trace-graph, attains the lowest cost among learning-based methods in almost every configuration under greedy decoding and in every configuration under sampling. Each baseline instead degrades in some regime: the MARL baselines exceed GPU memory on large MACTP instances, PARCO's success collapses on the largest MACTP and DFJSP instances, and VDN, DCG, and MACSIM leave a large fraction of DFJSP episodes incomplete under greedy decoding. The advantage of PACT grows under zero-shot transfer and is most pronounced on the largest MACTP configuration, where it reduces greedy cost by 34\% relative to MACSIM while maintaining 99.6\% success. Compared with online replanning, PACT reaches comparable or lower cost with sampling and runs at least seven times faster under greedy decoding wherever replanning completes, whereas replanning loses success or exceeds its time limit as instances grow. We note that, while \textit{Oracle} methods are not directly comparable due to having privileged access to the uncertainty $\xi$, they nevertheless may not yield optimal solutions on the largest instances, since they are limited by the time budget. We provide a breakdown of PACT's end-to-end wall clock runtime into environment stepping, encoding, candidate plan generation, coordination, plan selection, and other costs in Appendix~\ref{app:runtime-analysis}.

Validation returns obtained during training are reported in Figure~\ref{fig:training}. On MACTP, PACT and the MARL baselines reach similar final returns. On DFJSP, PACT reaches high-return performance earlier than MAPPO and maintains it thereafter, with lower variability across seeds than VDN and DCG, whose returns deteriorate after initial gains. On NCFFSP, PACT also learns faster than MAPPO and attains a higher final mean return. We present PARCO and MACSIM's training curves separately in Appendix~\ref{app:training-curves} because they require substantially more training steps. Despite larger budgets, PARCO attains lower best mean returns than PACT on all three problems, while MACSIM converges to a comparable return on MACTP and DFJSP but remains lower on NCFFSP.

\begin{figure}[t]
    \centering
    \includegraphics[width=\linewidth]{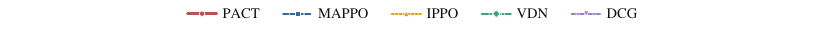}

    \begin{subfigure}[b]{0.32\linewidth}
        \centering
        \includegraphics[width=\linewidth]{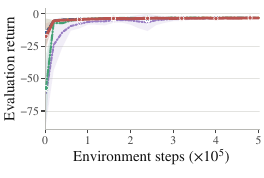}
        \caption{MACTP}
        \label{fig:train-mactp}
    \end{subfigure}
    \hfill
    \begin{subfigure}[b]{0.32\linewidth}
        \centering
        \includegraphics[width=\linewidth]{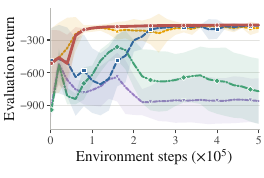}
        \caption{DFJSP}
        \label{fig:train-dfjsp}
    \end{subfigure}
    \hfill
    \begin{subfigure}[b]{0.32\linewidth}
        \centering
        \includegraphics[width=\linewidth]{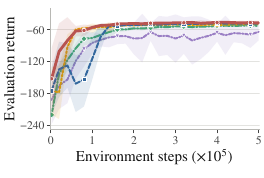}
        \caption{NCFFSP}
        \label{fig:train-ncffsp}
    \end{subfigure}

    \caption{Validation return during training. Curves show the mean across ten seeds; shaded regions indicate one standard deviation. Higher is better.}
    \label{fig:training}
\end{figure}

\subsection{Ablation Study}
\label{sec:ablation}

We ablate PACT on the largest zero-shot transfer configuration of each problem under greedy evaluation and report cost and runtime relative to the reference model with $K=4$ candidate plans and $R=2$ coordination rounds (Figure~\ref{fig:ablation}). Appendix~\ref{app:ablation} details the protocol and the analysis of each component.
Coordination is the main source of PACT's gains. Removing message passing on the plan-augmented trace-graph degrades performance to below that of MAPPO on DFJSP and NCFFSP, so candidate generation alone does not explain the improvement. Coordination also depends on having alternatives to choose from. With a single candidate plan per agent, performance degrades on every problem, whereas four candidates provide most of the benefit and larger sets yield no consistent gain. A second coordination round matters on MACTP, where plans span $H_p=4$ steps and capture multi-step interactions, while one round suffices on the scheduling problems with $H_p=1$. Further rounds bring no consistent improvement at nearly unchanged runtime.

\begin{figure}[tbh]
\centering
\begin{subfigure}[b]{0.32\linewidth}
    \centering
    \includegraphics[width=\linewidth]{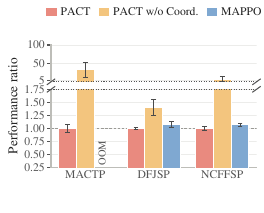}
    \caption{Effect of Coordination}
    \label{fig:ablation-graph}
\end{subfigure}
\hfill
\begin{subfigure}[b]{0.32\linewidth}
    \centering
    \includegraphics[width=\linewidth]{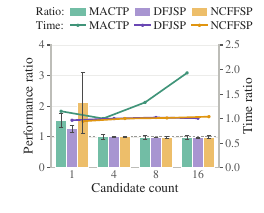}
    \caption{Number of Candidate Plans}
    \label{fig:ablation-k}
\end{subfigure}
\hfill
\begin{subfigure}[b]{0.32\linewidth}
    \centering
    \includegraphics[width=\linewidth]{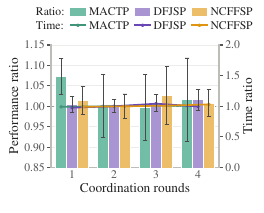}
    \caption{Coordination Rounds}
    \label{fig:ablation-rounds}
\end{subfigure}

\caption{Analysis of PACT components.}
\label{fig:ablation}
\end{figure}

\FloatBarrier
\endgroup

\section{Conclusion}
\label{sec:conclusion}

In this paper, we introduced PACT, a framework for DMACO that enables agents to coordinate over candidate plans through the environmental structure that induces their dependencies. By augmenting the environment graph with candidate plan traces and performing capacity-aware message passing over the resulting trace-graph, PACT extends coordination beyond the current decision step while retaining closed-loop execution as new information is revealed. Our experiments on routing and scheduling problems demonstrate that PACT consistently achieves strong solution quality and completion success across both in-distribution and zero-shot transfer settings, with particularly strong generalization to larger instances. In terms of limitations, the global communication assumption behind MPOMDPs may be relaxed in future work by imposing communication radiuses inside which information is shared. Additionally, PACT samples independently from agents' policies $\pi_i$ to construct the joint action at inference time. Stronger joint actions may be obtained by reasoning over the policies of other agents, for example by maintaining and updating beliefs. 

\endgroup

\label{lastmainpage}

\clearpage

\subsection*{AI use statement}
In this work, we used generative AI tools to design or provide feedback on research methodology or experiments, and to implement methods.
We have not used generative AI tools to
generate synthetic data sets,
help develop theoretical models or conceptual frameworks,
formulate mathematical claims,
provide critical ingredients for proving mathematical claims,
assist in the writing of proofs,
propose or refine hypotheses, or
interpret results.
Cleaning and reformatting datasets and
supporting qualitative and thematic data analysis
are not applicable to this work. 
Additionally, we used generative AI tools to 
draft parts of the research paper,
discover research topics or identify gaps, brainstorm, source/search for information, edit the research paper to improve readability, identify relevant literature, and assist with translation.
We have reviewed all AI-assisted work.
We checked all sources and LLM-generated research ideas for potential plagiarism through a manual literature review. All LLM-generated code was validated and tested for correctness.
All LLM-generated writing has been manually reviewed for accuracy.
We take full responsibility for the final content of this work, including text, claims, and artifacts produced with the aid of generative AI.

\subsection*{Reproducibility Statement}
Source code used to implement the PACT method will be released upon publication of the paper. Appendix~\ref{app:problem-defs} gives the formal model and revelation rules of each problem, and Appendix~\ref{app:datagen} specifies the instance generators. Appendix~\ref{app:baselines} describes the baseline implementations, and Appendix~\ref{app:hyperparams} and Appendix~\ref{app:hardware} report the training configurations and the compute environment. Appendix~\ref{app:algorithms} details the algorithms of PACT.

\bibliography{refs}
\bibliographystyle{iclr2027_conference}

\appendix
\newpage
\section{Problem Definitions}
\label{app:problem-defs}

This appendix defines the optimization problems and residual deterministic models (i.e., models of the remaining problem at each decision step) used by the baselines. 

\subsection{MACTP}
\label{app:mactp}

\paragraph{Problem Definition.} MACTP generalizes the Canadian Traveler Problem \citep{papadimitriou1991shortest} to a cooperative team that starts from distinct nodes and covers shared goals on a connected undirected graph with positive traversal costs.
Agents try to reach each goal by traversing a graph, but a subset of edges may be blocked, with a given probability, and the true state of the edge is only revealed when an agent reaches one of its endpoints.
The blockage probability of each edge is determined at instance generation time.
In order to fulfill a goal, an agent must take the action to claim it, which retires that agent.
Agents may pass through a goal without claiming it, and may pass through an already claimed goal.
The episode succeeds once all goals have been claimed.
A representative MACTP example was shown in Figure~\ref{fig:motivation} in the main text.

We formulate the optimization problem as finding a minimum-cost collection of walks
that jointly cover the remaining goals.

\paragraph{Indices and Sets}
\begin{itemize}[nosep]
    \item $\mathcal{N}=\{1,\ldots,n\}$ denotes the nodes, and
    $\mathcal{E}\subseteq\mathcal{N}\times\mathcal{N}$ contains the directed arcs induced by the
    usable undirected edges, with cost $w_{vv'}>0$.
    \item $\mathcal{D} \subseteq \mathcal{N}$: the set of goal nodes.
    \item $i\in\mathcal{M}$: agents represented in the current planning problem, with agent
    $i$ starting its residual walk at $s_i$.
    \item $\mathcal{D}^{\mathrm{cov}}\subseteq\mathcal{D}$: goals still requiring coverage,
    with $G_t=|\mathcal{D}^{\mathrm{cov}}|$.
\end{itemize}

\paragraph{Decision Variables}
\begin{itemize}[nosep]
    \item $x_{i,(v,v')}\in\mathbb{Z}_{\ge0}$: number of traversals of arc $(v,v')$ by
    agent $i$.
    \item $f_{i,(v,v')}\ge0$: auxiliary connectivity flow carried only on arcs used by
    agent $i$.
    \item $y_{i,v}\in\{0,1\}$ indicates whether agent $i$ is credited with covering $v$ and
    is fixed to zero for $v\notin\mathcal{D}^{\mathrm{cov}}$.
    \item $e_{i,v}\in\{0,1\}$: whether agent $i$ ends its residual walk at
    $v\in\mathcal{N}$.
\end{itemize}

\paragraph{Objective Function}
\begin{small}
\begin{align}
    \min \quad \sum_{i\in\mathcal{M}}\sum_{(v,v')\in\mathcal{E}}
    w_{vv'}x_{i,(v,v')} \label{eq:mactp-obj}
\end{align}
\end{small}

\paragraph{Subject to}
\begin{small}
\begin{align}
    & \sum_{v\in\mathcal{N}}e_{i,v}=1,
        && \forall i\in\mathcal{M} \label{eq:mactp-end} \\
    & \sum_{(v,v')\in\mathcal{E}}x_{i,(v,v')}
        -\sum_{(v',v)\in\mathcal{E}}x_{i,(v',v)}
        =\mathbf{1}[v=s_i]-e_{i,v},
        && \forall i\in\mathcal{M},\ v\in\mathcal{N} \label{eq:mactp-degree} \\
    & \sum_{(v,v')\in\mathcal{E}}f_{i,(v,v')}
        -\sum_{(v',v)\in\mathcal{E}}f_{i,(v',v)}
        =\Big(\sum_{g\in\mathcal{D}^{\mathrm{cov}}}y_{i,g}\Big)\mathbf{1}[v=s_i]
        -y_{i,v}\mathbf{1}[v\in\mathcal{D}^{\mathrm{cov}}],
        && \forall i\in\mathcal{M},\ v\in\mathcal{N} \label{eq:mactp-flow} \\
    & f_{i,(v,v')}\le G_t x_{i,(v,v')},
        && \forall i\in\mathcal{M},\ (v,v')\in\mathcal{E} \label{eq:mactp-cap} \\
    & \sum_{i\in\mathcal{M}}y_{i,v}=1,
        && \forall v\in\mathcal{D}^{\mathrm{cov}} \label{eq:mactp-cover} 
\end{align}
\end{small}

\paragraph{Constraints and Online Replanning.} The degree constraints define one walk per agent, while the auxiliary flow certifies that every credited goal lies on that walk. 
During online replanning, revealed blocked edges are removed, whereas unrevealed edges remain available under their optimistic status. 
If the revealed graph makes complete coverage infeasible, we replace Eq.~\eqref{eq:mactp-cover} with $\sum_i y_{i,v}+u_v=1$ and lexicographically minimize the number of uncovered goals before travel cost.

\paragraph{MPOMDP Formulation.} In the notation of \Cref{sec:preliminaries}, the latent variable $\xi$ defines the blockage probability of the edges, while the graph state $\mathcal{G}_t$ is the known graph state.
Node features contain goal coverage and claim status. 
Edge features contain traversal cost, blocking probability, and the currently known status, while each agent carries its present location as an agent feature.

The active set $\mathcal{M}_t$ contains the whole team, although retired agents can only remain at their current nodes and live agents move only through deterministic or revealed-open edges.
An agent may retire by claiming an unclaimed goal, where unit capacity and unit claim demand enforce exclusivity, while non-goal nodes impose no shared constraint.

The step cost $C(\mathcal{G}_t,\mathbf{a}_t,\xi)$ is the team's total traversal cost, with retired agents incurring zero cost. Episodes end after all goals are covered or all agents retire, and truncated episodes additionally receive an estimate of the remaining completion cost.

\subsection{DFJSP}
\label{app:dfjsp}

\begin{figure}[h]
    \centering
    \includegraphics[width=\linewidth]{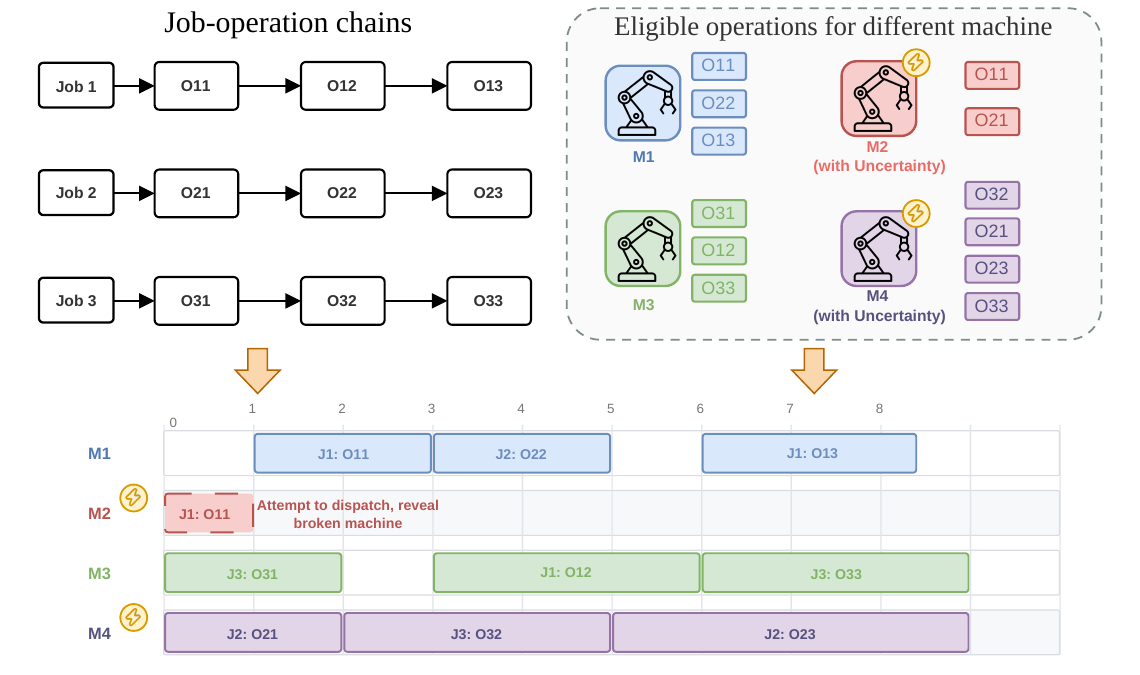}
    \caption{Illustration of the DFJSP task. The upper panel pairs ordered job chains with machine-specific eligibility sets, where the lightning bolts mark machines that may have latent breakdowns. 
    The lower panel gives a machine-wise schedule, where the dashed block records a failed dispatch to a broken machine and the solid block shows successful operation dispatch.}
    \label{fig:app-dfjsp-scenario}
\end{figure}

\paragraph{Problem Definition.} In DFJSP, every job follows an ordered chain of operations, and each operation is eligible for a specified subset of machines. 
Some machines are unavailable from the beginning of an episode, but their condition remains hidden until the first attempted dispatch.
A failed dispatch leaves the operation unscheduled and delays its job by a fixed amount, after which the broken machine is removed from future decisions.
Failures are correlated through latent groups whose memberships and priors are observed, while the group condition and each machine-specific outcome remain hidden.
Machines act as agents and may schedule operations before becoming idle, with processing beginning once both the selected machine and job are available. The objective function is the makespan of the resulting schedule.
Figure~\ref{fig:app-dfjsp-scenario} visualizes an illustrative DFJSP instance.

We formulate the optimization problem as a standard disjunctive FJSP over the remaining operations and the machines currently considered usable \citep{ozguven2010mathematical}.

\paragraph{Indices and Sets}
\begin{itemize}
    \item $j\in\mathcal{J}$ indexes jobs, $\mathrm{chain}(j)$ is the ordered operation chain
    of job $j$, and $\mathcal{O}_t$ contains the operations not yet scheduled at replanning step
    $t$.
    \item $\mathrm{Elig}_t(o)\subseteq\mathcal{I}$: machines currently considered usable
    for operation $o$, with processing time $p_{o,i}>0$ for $i\in\mathrm{Elig}_t(o)$.
    \item $r_{o,t}$ is the earliest time permitted by the current job state, and $q_{i,t}$
    is the next time at which machine $i$ becomes available.
\end{itemize}

\paragraph{Decision Variables}
\begin{itemize}
    \item $x_{o,i}\in\{0,1\}$: whether operation $o$ is assigned to machine
    $i\in\mathrm{Elig}_t(o)$.
    \item $s_o\ge0$ is the start time of operation $o$, and
    $d_o=\sum_{i\in\mathrm{Elig}_t(o)}p_{o,i}x_{o,i}$ is its assigned duration.
    \item $z_{o,o'}\in\{0,1\}$: relative order of two distinct operations that may share
    a machine, defined once for each unordered pair $o<o'$.
    \item $C_{\max}\ge0$ is the makespan, and $B$ is any valid upper bound on the completion
    time of the residual schedule.
\end{itemize}

\paragraph{Objective Function}
\begin{small}
\begin{align}
    \min \quad C_{\max} \label{eq:sched-obj}
\end{align}
\end{small}

\paragraph{Subject to}
\begin{small}
\begin{align}
    & \sum_{i\in\mathrm{Elig}_t(o)}x_{o,i}=1,
        && \forall o\in\mathcal{O}_t \label{eq:sched-assign} \\
    & C_{\max}\ge s_o+d_o,
        && \forall o\in\mathcal{O}_t \label{eq:sched-cmax} \\
    & s_{o'}\ge s_o+d_o,
        && \forall (o,o')\text{ consecutive in a residual job chain} \label{eq:sched-prec} \\
    & s_o\ge r_{o,t},
        && \forall o\in\mathcal{O}_t \label{eq:sched-ready} \\
    & s_o\ge q_{i,t}-B(1-x_{o,i}),
        && \forall o\in\mathcal{O}_t,\ i\in\mathrm{Elig}_t(o) \label{eq:sched-release} \\
    & s_{o'}\ge s_o+d_o-B(1-z_{o,o'})-B(2-x_{o,i}-x_{o',i}),
        && \forall o<o',\ i\in\mathrm{Elig}_t(o)\cap\mathrm{Elig}_t(o') \label{eq:sched-disj1} \\
    & s_o\ge s_{o'}+d_{o'}-Bz_{o,o'}-B(2-x_{o,i}-x_{o',i}),
        && \forall o<o',\ i\in\mathrm{Elig}_t(o)\cap\mathrm{Elig}_t(o') \label{eq:sched-disj2}
\end{align}
\end{small}

\paragraph{Constraints and Online Replanning.} Each remaining operation is assigned once and must follow its job predecessor, while the release constraints preserve the current job and machine timelines.
The pairwise disjunctions prevent operations on the same machine from overlapping. 
At each replanning step, completed operations and revealed-broken machines are removed, whereas unrevealed machines remain provisionally available with their known processing times.

\paragraph{MPOMDP Formulation.} The latent variable $\xi$ records the condition of every uncertain machine, with correlations induced by group-level variables.
Each element of $\mathcal{G}_t$ represents a job and records its next unscheduled operation and earliest feasible start time. 
Machine features contain availability, group membership, and the current condition estimate, while an eligibility edge carries the processing time for the job's next operation.

The active set $\mathcal{M}_t$ contains every machine not known to be broken, including busy machines, which may either dispatch an eligible job or wait. 
A scheduled operation starts no earlier than the release times of both participants, while unit job capacity permits at most one dispatch to each job per round.

The step cost $C(\mathcal{G}_t,\mathbf{a}_t,\xi)$ is the increase in makespan, so its undiscounted sum equals $C_{\max}$ for a completed schedule.
A truncated episode additionally receives a penalty of processing all remaining operations serially on their slowest available machines, providing a pessimistic completion cost that prevents stalling from being beneficial.

\subsection{NCFFSP}
\label{app:ncffsp}

\begin{figure}[H]
    \centering
    \includegraphics[width=\linewidth]{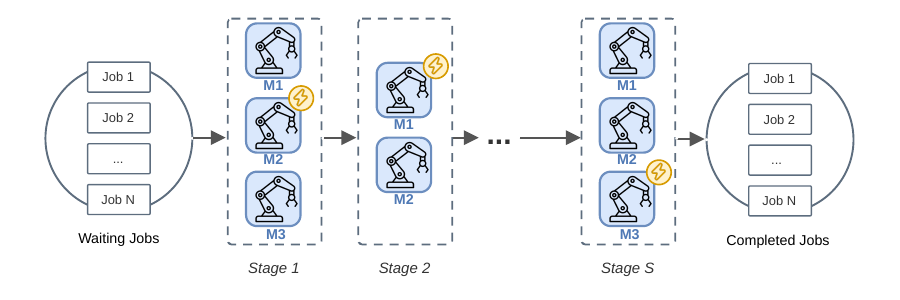}
    \caption{Illustration of the NCFFSP task. Jobs operate from left to right through consecutive parallel machine stages, where the lightning bolts mark machines that may have unrevealed duration factors.}
    \label{fig:app-ncffsp-scenario}
\end{figure}

\paragraph{Problem Definition.} In NCFFSP, jobs pass through a common sequence of stages, and each stage is served by parallel machines whose processing times depend on both the job and the selected machine \citep{kwon2021matnet}.
Some machines also have hidden condition factors that scale their processing times and are correlated through latent batches. 
Batch membership and its prior are observed, but a machine's realized factor appears only after its first dispatch.
Unlike DFJSP, every dispatch succeeds because uncertainty changes processing duration rather than machine eligibility, and busy machines may schedule jobs that begin once both participants are available.
The objective is the final makespan, and non-clairvoyance here refers to the initially unknown machine-dependent duration factors \citep{motwani1994nonclairvoyant}.
Figure~\ref{fig:app-ncffsp-scenario} illustrates an NCFFSP workflow.

We formulate the optimization problem over the remaining job stages as a flexible flow-shop model using the current processing times.
Assignment, stage-precedence, release-time, and machine non-overlap constraints are used.

\paragraph{Indices and Sets}
\begin{itemize}
    \item $\ell\in\{1,\ldots,S\}$ indexes stages, and $\mathcal{I}_\ell$ contains the
    machines serving stage $\ell$.
    \item $j,j'\in\mathcal{J}_t$: jobs with at least one stage still to be scheduled in
    the residual problem.
    \item $\bar p_{\ell ji}>0$: nominal processing time of job $j$ on machine
    $i\in\mathcal{I}_\ell$.
    \item $\xi_{\ell i}$: hidden condition factor of machine $i$. Before revelation,
    $p_{\ell ji,t}=\bar p_{\ell ji}$, while afterwards it is the positive rounded value of
    $\bar p_{\ell ji}\xi_{\ell i}$.
    \item $r_{\ell j,t}$ is the current release-time floor for job $j$ at stage $\ell$, and
    $q_{i,t}$ is the next time at which machine $i$ becomes available.
\end{itemize}

\paragraph{Decision Variables}
\begin{itemize}
    \item $x_{\ell ji}\in\{0,1\}$: whether job $j$ is assigned to machine
    $i\in\mathcal{I}_\ell$ at stage $\ell$.
    \item $s_{\ell j}\ge0$ is the start time, and
    $d_{\ell j}=\sum_{i\in\mathcal{I}_\ell}p_{\ell ji,t}x_{\ell ji}$ is the assigned duration.
    \item $z_{\ell jj'}\in\{0,1\}$: relative order of distinct jobs $j<j'$ that may
    share a machine at stage $\ell$.
    \item $C_{\max}\ge0$ is the makespan, and $B$ is any valid upper bound on the completion
    time of the residual schedule.
\end{itemize}

\paragraph{Objective Function}
\begin{small}
\begin{align}
    \min \quad C_{\max} \label{eq:ncffsp-obj}
\end{align}
\end{small}

\paragraph{Subject to}
\begin{small}
\begin{align}
    & \sum_{i\in\mathcal{I}_\ell}x_{\ell ji}=1,
        && \forall j,\ \ell\text{ remaining for }j \label{eq:ncffsp-assign} \\
    & C_{\max}\ge s_{\ell j}+d_{\ell j},
        && \forall j,\ \ell\text{ remaining for }j \label{eq:ncffsp-cmax} \\
    & s_{\ell+1,j}\ge s_{\ell j}+d_{\ell j},
        && \forall j,\ \ell<S\text{ remaining for }j \label{eq:ncffsp-prec} \\
    & s_{\ell j}\ge r_{\ell j,t},
        && \forall j,\ \ell\text{ remaining for }j \label{eq:ncffsp-ready} \\
    & s_{\ell j}\ge q_{i,t}-B(1-x_{\ell ji}),
        && \forall j,\ \ell\text{ remaining for }j,\ i\in\mathcal{I}_\ell
        \label{eq:ncffsp-release} \\
    & s_{\ell j'}\ge s_{\ell j}+d_{\ell j}-B(1-z_{\ell jj'})
        -B(2-x_{\ell ji}-x_{\ell j'i}),
        && \forall \ell\text{ remaining for }j,j',\ j<j',\ i\in\mathcal{I}_\ell
        \label{eq:ncffsp-disj1} \\
    & s_{\ell j}\ge s_{\ell j'}+d_{\ell j'}-Bz_{\ell jj'}
        -B(2-x_{\ell ji}-x_{\ell j'i}),
        && \forall \ell\text{ remaining for }j,j',\ j<j',\ i\in\mathcal{I}_\ell
        \label{eq:ncffsp-disj2}
\end{align}
\end{small}

\paragraph{Constraints and Online Replanning} The assignment and non-overlap constraints
play the same roles as in DFJSP, while Eq.~\eqref{eq:ncffsp-prec} enforces the shared stage order.
The release bounds preserve the current job and machine timelines, while online replanning uses
nominal durations until a machine is revealed and realized durations thereafter. Every machine
remains eligible for future assignments.

\paragraph{MPOMDP Formulation.} NCFFSP uses the job-machine graph interface of DFJSP, but its latent variable $\xi$ contains machine-dependent duration factors rather than binary failure states, and each job element records its next stage and earliest start time. Machine features store the served stage, current availability, batch membership, and either the prior or the revealed factor, while each eligibility edge carries the effective processing time known at that step.

Every machine remains in $\mathcal{M}_t$, including those that are currently busy, and may
either dispatch a job at its stage or wait. Each job has unit capacity within a round, and a
dispatch directed to that job contributes unit demand, while the scheduled start time respects
both participants' release times and prevents assignments on the same machine from overlapping.

The step cost $C(\mathcal{G}_t,\mathbf{a}_t,\xi)$ is the increase in makespan, whose undiscounted sum equals the final makespan for a completed schedule.
A truncated episode additionally receives a penalty of the makespan required to complete the schedule from the truncated state under a uniformly random legal policy. This penalty is separate from the NCFFSP objective and ensures that giving up is no better than acting randomly.

\section{Experimental Details}
\label{app:baselines}

This appendix lists the baselines evaluated in \Cref{sec:experiments} and the licenses of their corresponding code implementations.
We also describe how instances are generated for each of the three problems, the training hyperparameters for every learning-based method, and the hardware and software used to run all experiments.

\subsection{Baselines}
\label{app:baseline-methods}

\paragraph{Classic Solvers.} The Gurobi and OR-Tools classic solvers are ran in two modes. The baselines marked \textit{Oracle} in the results solve complete realizations using privileged access to the uncertainty with a time limit of 10 minutes. The \textit{Replan} baselines repeatedly solve the remaining problem under the information available at each decision step, with a time limit of 2 seconds for each re-solve.

\paragraph{Gurobi.} Gurobi \citep{gurobi} solves the mixed-integer linear programs in Appendix~\ref{app:problem-defs} using branch-and-cut.
We retain the best feasible solution found for each instance under the time budget and report the mean objective value across the evaluation set.

\paragraph{OR-Tools.} OR-Tools \citep{ortools} provides a complementary CP-SAT implementation under the same information settings and evaluation configurations. 

\paragraph{Graph-based MARL Baselines.} MAPPO, IPPO, VDN, and DCG use matched graph encoders
over the common observation interface, and their per-agent networks share parameters within the
team while the flow of information among agents follows the centralized or decentralized
structure of each algorithm.

\paragraph{MAPPO.} Multi-Agent PPO \citep{yu2022mappo} uses a shared-parameter decentralized actor and a centralized critic, so each actor conditions on the common graph information and its own agent features while the critic aggregates information across the team. Both networks are trained with the clipped PPO objective and generalized advantage estimation.

\paragraph{IPPO.} Independent PPO \citep{dewitt2020ippo} uses the same shared-parameter decentralized actor as MAPPO, but replaces the centralized critic with a decentralized value function. 
The critic produces one estimate per agent from that agent's observation, and training uses the same clipped PPO objective and advantage estimator.

\paragraph{VDN.} Value Decomposition Networks \citep{sunehag2018vdn} learn shared per-agent action-value functions and select each agent's action independently, after which the selected utilities are summed into a team value, $Q_{\mathrm{tot}}=\sum_i Q_i$, trained from a common temporal-difference target.

\paragraph{DCG.} Deep Coordination Graphs \citep{bohmer2020dcg} combine per-agent utility terms with pairwise payoff terms defined on a fixed, fully connected coordination graph, and at each step max-plus message passing approximately maximizes the resulting joint action-value function to select the agents' actions.

\paragraph{PARCO.} PARCO \citep{berto2025parco} is a parallel autoregressive neural solver in which a shared transformer encodes the instance and communication layers exchange context among agent queries. A multiple-pointer decoder proposes simultaneous actions before a learned priority rule resolves conflicts and assigns fallback actions to displaced agents.

\paragraph{MACSIM.} MACSIM \citep{luttmann2026macsim} also constructs multi-agent solutions autoregressively from a shared instance representation. To produce a conflict-free set of simultaneous actions, it samples agent-task pairs without
replacement from a joint categorical distribution at each construction step.

\paragraph{Neural Solver Adaptation.} We implement task wrappers for PARCO and MACSIM through their respective environment interfaces while preserving the instances, transition rules, action constraints, and objectives used to evaluate the other methods.

\paragraph{Solution Decoding Strategies.} We evaluate PARCO, MACSIM, and PACT under two solution decoding strategies: \textit{greedy} and \textit{sampling}.
Greedy decoding uses one deterministic rollout, while sampling decoding generates $128$ independent stochastic rollouts per method and instance and retains that method's best realized solution \citep{berto2025parco,luttmann2026macsim}.
For greedy decoding, PARCO and MACSIM select the argmax of their action distributions, while in PACT the internal candidate plan generation, virtual rollout, and plan selection are all made deterministic.

\subsection{Code and Licenses}
\label{app:licenses}

Table~\ref{tab:licenses} lists the external solvers and codebases used to implement the
baselines above, together with their licenses.

\begin{table}[t]
\centering
\caption{Summary of licenses for baselines.}
\label{tab:licenses}
\resizebox{\textwidth}{!}{%
\begin{tabular}{lll}
\toprule
Resource & Type & License \\
\midrule
Gurobi \citep{gurobi}                       & Solver (commercial)          & Commercial, academic license \\
OR-Tools \citep{ortools}                    & Solver (open-source library) & Apache License 2.0 \\
BenchMARL \citep{bettini2024benchmarl} & Library (MAPPO, IPPO, and VDN) & MIT License \\
DCG \citep{bohmer2020dcg} & Code & Apache License 2.0 \\
PARCO \citep{berto2025parco}          & Code                    & MIT License \\
MACSIM \citep{luttmann2026macsim}     & Code                    & Apache License 2.0 \\
\bottomrule
\end{tabular}}
\end{table}

\subsection{Data Generation}
\label{app:datagen}

\paragraph{MACTP.} Following SALSA-CLRS \citep{minder2023salsa}, each instance uses a sparse, connected Erd\H{o}s--R\'enyi graph over $n$ nodes, with agent starts and goals sampled uniformly at random and positive traversal costs assigned to its edges.
In the reference configuration, 25\% of the edges are stochastic, and their traversability is sampled at generation time but remains hidden until encountered, as described in Appendix~\ref{app:mactp}.

\paragraph{DFJSP.} We follow the synthetic FJSP generation procedure of \citet{song2023flexible}: each job contains four to six ordered operations, each operation is assigned a random subset of eligible machines, and its machine-dependent processing times vary within 20\% of an operation-level mean.
Each machine is initially designated stochastic with probability $0.6$, after which a feasibility repair ensures that every operation retains at least one always-available eligible machine.
Stochastic machines have fixed hidden availability and correlated breakdown probabilities within small machine groups, so observing one machine can inform the condition of others. 
Availability is revealed only when a job is first dispatched to the machine, and the resulting transition follows Appendix~\ref{app:dfjsp}.

\paragraph{NCFFSP.} Each instance contains $N$ jobs that pass through $S$ processing stages, with several interchangeable machines available at each stage. 
Following \citet{kwon2021matnet}, nominal job--machine processing times are sampled independently and uniformly from the integers $\{2,\ldots,9\}$. Each real machine is independently designated stochastic with probability $0.6$ and receives a fixed hidden condition factor generated from a group-correlated latent variable $u\in[0,1]$ through $\xi=0.8+7.2u^4$.
The factor is revealed on the machine's first dispatch and scales its processing times thereafter, as described in Appendix~\ref{app:ncffsp}.

\subsection{Training Configuration}
\label{app:hyperparams}

PACT and the four MARL baselines use a common number of training steps, data-collection scale, optimizer, and graph-encoder configuration across all three tasks. 
Table~\ref{tab:marl-training} reports these shared settings together with the update parameters specific to the on-policy and value-based families. 
We evaluate checkpoints on the same 64 evaluation episodes every 20{,}000 steps, and each configuration is trained with 10 random seeds.

\begin{table}[t]
\centering
\caption{Training configuration for PACT and the MARL baselines.}
\label{tab:marl-training}
\begin{tabular}{@{}ll@{}}
\toprule
Hyperparameter & Value \\
\midrule
\multicolumn{2}{@{}l}{\textit{Shared settings}} \\
Training steps & 500{,}000 \\
Validation interval & 20{,}000 steps \\
Validation episodes & 64 \\
Random seeds & 10 \\
Optimizer & Adam \\
Learning rate & $5\times10^{-5}$ \\
Steps per batch & 2{,}000 \\
Parallel environments & 4 \\
Maximum gradient norm & 5 \\
Encoder layers & 2 \\
Hidden dimension & 64 \\
Aggregation & Weighted mean \\
Discount factor & $0.99$ (MACTP); $1.0$ (DFJSP and NCFFSP) \\
\midrule
\multicolumn{2}{@{}l}{\textit{PPO settings (PACT, MAPPO, and IPPO)}} \\
Update epochs & 4 \\
Minibatch size & 128 \\
Clipping coefficient & $0.2$ \\
GAE parameter $\lambda$ & $0.95$ \\
Entropy coefficient (MAPPO and IPPO) & $0.01$ \\
Value-loss coefficient & $1.0$ \\
\midrule
\multicolumn{2}{@{}l}{\textit{Value-based settings (VDN and DCG)}} \\
Updates per batch & 200 \\
Minibatch size & 256 \\
Replay-buffer capacity & 30{,}000 \\
Target-update coefficient & $0.005$ \\
Initial exploration rate $\epsilon$ & $0.8$ \\
\midrule
\multicolumn{2}{@{}l}{\textit{DCG-specific settings}} \\
Payoff rank & 8 \\
Max-plus iterations & 4 \\
\midrule
\multicolumn{2}{@{}l}{\textit{PACT-specific settings}} \\
Score step bound $\eta$ & $1.0$ (fixed) \\
Selection temperature $\tau$ & Learned, initialized to $1.0$, $\tau \ge 0.25$ \\
Coordination MPNN layers & $1$ \\
Coordination hidden dimension & $64$ \\
\bottomrule
\end{tabular}
\end{table}

\paragraph{PACT-specific Settings.} PACT uses two coordination rounds in every task and is evaluated with $K=4$ candidate plans per agent. During training, $K$ is sampled from $\{2,4,8\}$, while the planning horizon is $H_p=4$ on MACTP and $H_p=1$ on DFJSP and NCFFSP.
All three decision groups use a clipping range of $0.2$. The proposal and rollout groups use an entropy coefficient of $0.001$, while the selection group uses $0.01$.

\paragraph{Neural Combinatorial Optimization Baselines.} PARCO and MACSIM retain their original training objectives and model configurations, with PARCO using its rollout-based objective and MACSIM using self-improvement.
Both methods use Adam with learning rate $1\times10^{-4}$ and are trained for 50 epochs on each task.

\FloatBarrier

\subsection{Compute and Software}
\label{app:hardware}

All learning-based methods were trained and evaluated on NVIDIA L40S
GPUs, while the Gurobi and OR-Tools baselines were run on an Intel Core Ultra 7 265HX CPU.
Experiments used Python 3.11,
PyTorch 2.9.1, TorchRL 0.10.1, TensorDict 0.10.0, PyTorch Geometric 2.6.1, NumPy 2.1.3, and
NetworkX 3.6.1. The optimization baselines used Gurobi 13.0.3 and OR-Tools 9.12.4544.

\section{Additional Experimental Results}
\label{app:additional-results}

\subsection{Training Curves for PARCO and MACSIM}
\label{app:training-curves}

\begin{figure}[htb]
    \centering
    \includegraphics[width=\linewidth]{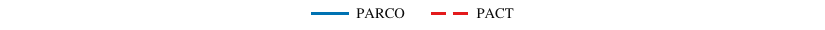}

    \begin{subfigure}[b]{0.32\linewidth}
        \centering
        \includegraphics[width=\linewidth]{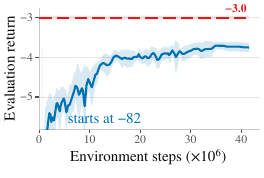}
        \caption{MACTP}
        \label{fig:app-train-parco-mactp}
    \end{subfigure}
    \hfill
    \begin{subfigure}[b]{0.32\linewidth}
        \centering
        \includegraphics[width=\linewidth]{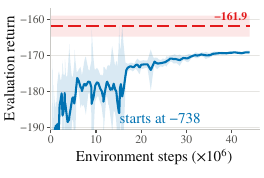}
        \caption{DFJSP}
        \label{fig:app-train-parco-dfjsp}
    \end{subfigure}
    \hfill
    \begin{subfigure}[b]{0.32\linewidth}
        \centering
        \includegraphics[width=\linewidth]{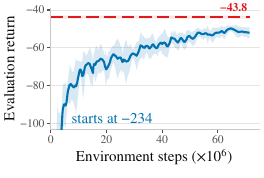}
        \caption{NCFFSP}
        \label{fig:app-train-parco-ncffsp}
    \end{subfigure}

    \caption{Validation return during training for PARCO on the three DMACO problems.}
    \label{fig:app-train-parco}
\end{figure}

\begin{figure}[htb]
    \centering
    \includegraphics[width=\linewidth]{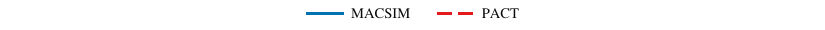}

    \begin{subfigure}[b]{0.32\linewidth}
        \centering
        \includegraphics[width=\linewidth]{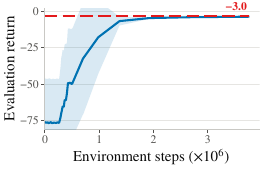}
        \caption{MACTP}
        \label{fig:app-train-macsim-mactp}
    \end{subfigure}
    \hfill
    \begin{subfigure}[b]{0.32\linewidth}
        \centering
        \includegraphics[width=\linewidth]{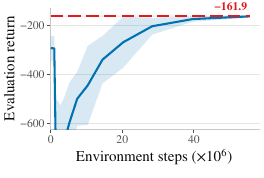}
        \caption{DFJSP}
        \label{fig:app-train-macsim-dfjsp}
    \end{subfigure}
    \hfill
    \begin{subfigure}[b]{0.32\linewidth}
        \centering
        \includegraphics[width=\linewidth]{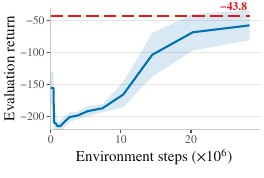}
        \caption{NCFFSP}
        \label{fig:app-train-macsim-ncffsp}
    \end{subfigure}

    \caption{Validation return during training for MACSIM on the three DMACO problems.}
    \label{fig:app-train-macsim}
\end{figure}

Figures~\ref{fig:app-train-parco} and~\ref{fig:app-train-macsim} show the validation returns of PARCO and MACSIM during training on the three DMACO problems.
Curves show the mean across ten seeds, with shaded regions indicating one standard deviation. The horizontal red dashed line marks PACT's highest mean validation return during training, and the red shaded band shows one standard deviation across seeds at the corresponding checkpoint. 
PACT achieves a higher best mean validation return than PARCO across all three problems and outperforms MACSIM on NCFFSP. 
While it converges to returns comparable to PACT on MACTP and DFJSP, MACSIM consistently requires substantially more training steps across all tasks. 
Overall, PACT demonstrates a significantly improved trade-off between sample efficiency and validation performance. Furthermore, despite occasional matching validation performance in-distribution, we nevertheless highlight the significantly higher performance on larger ZST instances reported in the main text.

\subsection{Complete Results for Learning-Based Methods}
\label{app:complete-results}

We evaluate each method on 100 instances for each of 10 independently trained seeds and report the mean across seeds. 
Success is the percentage of successfully completed instances, cost is the mean negated episodic return over all episodes, including penalties from failed or truncated episodes, whereas objective is the mean negated episodic return over successfully completed episodes.
Performance ratio is cost divided by the mean cost of PACT on the same problem and test configuration within the same evaluation group: greedy methods use greedy PACT as the reference, and sampling methods use sampling PACT.
PACT has a ratio of one within each group, and lower values are better. All ratios are computed from the cost values as reported in the table. 
Time reports the end-to-end runtime per instance in seconds. Each entry reports the mean and one standard deviation across the ten training seeds; performance ratio is computed from the mean cost and reported without a standard deviation.
N/A indicates that the method does not support the configuration.

Tables~\ref{tab:complete-mactp}, \ref{tab:complete-dfjsp}, and~\ref{tab:complete-ncffsp} report these metrics under greedy and best-of-128 sampling evaluation.

\newlength{\completeTableColumnWidth}

{\small
\setlength{\aboverulesep}{0pt}
\setlength{\belowrulesep}{0pt}
\renewcommand{\arraystretch}{1.0}
\setlength{\tabcolsep}{3pt}
\setlength{\completeTableColumnWidth}{\dimexpr0.98\textwidth-12\tabcolsep-6\arrayrulewidth\relax}
\begin{longtable}{@{}>{\raggedright\arraybackslash}m{0.11\completeTableColumnWidth}|>{\raggedright\arraybackslash}m{0.12\completeTableColumnWidth}|>{\centering\arraybackslash}m{0.14\completeTableColumnWidth}|>{\centering\arraybackslash}m{0.17\completeTableColumnWidth}|>{\centering\arraybackslash}m{0.17\completeTableColumnWidth}|>{\centering\arraybackslash}m{0.12\completeTableColumnWidth}|>{\centering\arraybackslash}m{0.17\completeTableColumnWidth}@{}}
\caption{Complete results for learning-based methods on MACTP. The
first configuration is in-distribution (ID), and the remaining configurations are evaluated
zero-shot (ZST).}
\label{tab:complete-mactp}\\
\toprule
Method & Evaluation & Success (\%) $\uparrow$ & Cost $\downarrow$ & Objective $\downarrow$ & Perf. ratio $\downarrow$ & Time (s) $\downarrow$ \\
\midrule
\endfirsthead
\multicolumn{7}{c}{\tablename~\thetable\ continued}\\
\toprule
Method & Evaluation & Success (\%) $\uparrow$ & Cost $\downarrow$ & Objective $\downarrow$ & Perf. ratio $\downarrow$ & Time (s) $\downarrow$ \\
\midrule
\endhead
\bottomrule
\endfoot
\multicolumn{7}{c}{\textbf{$16\times4\times6$} (ID)}\\*
\noalign{\nobreak\hrule height\lightrulewidth\nobreak}
MAPPO  & Greedy & 96.8{\scriptsize$\pm$2.1} & 3.83{\scriptsize$\pm$0.25} & 3.53{\scriptsize$\pm$0.06} & 1.046 & 0.027{\scriptsize$\pm$0.030} \\*
IPPO   & Greedy & 97.1{\scriptsize$\pm$2.0} & 3.92{\scriptsize$\pm$0.36} & 3.57{\scriptsize$\pm$0.11} & 1.071 & 0.015{\scriptsize$\pm$0.007} \\*
VDN    & Greedy & 98.9{\scriptsize$\pm$0.9} & \textbf{3.52}{\scriptsize$\pm$0.20} & \textbf{3.33}{\scriptsize$\pm$0.04} & 0.962 & 0.012{\scriptsize$\pm$0.004} \\*
DCG    & Greedy & \textbf{99.3}{\scriptsize$\pm$0.7} & 3.68{\scriptsize$\pm$0.47} & 3.55{\scriptsize$\pm$0.26} & 1.005 & 0.011{\scriptsize$\pm$0.003} \\*
PARCO  & Greedy & 98.9{\scriptsize$\pm$1.2} & 4.56{\scriptsize$\pm$0.23} & 4.34{\scriptsize$\pm$0.13} & 1.246 & 0.004{\scriptsize$\pm$0.002} \\*
MACSIM & Greedy & 89.9{\scriptsize$\pm$3.7} & 7.36{\scriptsize$\pm$0.68} & 5.55{\scriptsize$\pm$0.35} & 2.011 & 0.038{\scriptsize$\pm$0.021} \\*
PACT   & Greedy & 99.1{\scriptsize$\pm$1.4} & 3.66{\scriptsize$\pm$0.18} & 3.43{\scriptsize$\pm$0.07} & 1.000 & 0.056{\scriptsize$\pm$0.015} \\*
\midrule
PARCO  & Sampling & \textbf{100.0}{\scriptsize$\pm$0.0} & 3.35{\scriptsize$\pm$0.04} & 3.35{\scriptsize$\pm$0.04} & 1.159 & 0.094{\scriptsize$\pm$0.016} \\*
MACSIM & Sampling & \textbf{100.0}{\scriptsize$\pm$0.0} & 3.27{\scriptsize$\pm$0.09} & 3.27{\scriptsize$\pm$0.09} & 1.131 & 0.48{\scriptsize$\pm$0.07} \\*
PACT   & Sampling & \textbf{100.0}{\scriptsize$\pm$0.0} & \textbf{2.89}{\scriptsize$\pm$0.02} & \textbf{2.89}{\scriptsize$\pm$0.02} & 1.000 & 5.27{\scriptsize$\pm$0.49} \\
\midrule
\multicolumn{7}{c}{\textbf{$32\times6\times10$} (ZST)}\\*
\noalign{\nobreak\hrule height\lightrulewidth\nobreak}
MAPPO  & Greedy & 98.1{\scriptsize$\pm$1.8} & 6.78{\scriptsize$\pm$0.68} & 6.19{\scriptsize$\pm$0.19} & 1.171 & 0.020{\scriptsize$\pm$0.007} \\*
IPPO   & Greedy & 98.8{\scriptsize$\pm$1.4} & 6.84{\scriptsize$\pm$0.76} & 6.27{\scriptsize$\pm$0.22} & 1.181 & 0.033{\scriptsize$\pm$0.052} \\*
VDN    & Greedy & \textbf{99.6}{\scriptsize$\pm$1.0} & 5.84{\scriptsize$\pm$0.40} & 5.69{\scriptsize$\pm$0.23} & 1.009 & 0.019{\scriptsize$\pm$0.019} \\*
DCG    & Greedy & 98.1{\scriptsize$\pm$1.4} & 8.44{\scriptsize$\pm$4.49} & 7.84{\scriptsize$\pm$4.36} & 1.458 & 0.048{\scriptsize$\pm$0.029} \\*
PARCO  & Greedy & 95.0{\scriptsize$\pm$2.4} & 10.8{\scriptsize$\pm$1.8} & 7.67{\scriptsize$\pm$0.25} & 1.865 & 0.017{\scriptsize$\pm$0.005} \\*
MACSIM & Greedy & 95.2{\scriptsize$\pm$5.1} & 12.7{\scriptsize$\pm$1.3} & 11.0{\scriptsize$\pm$0.7} & 2.193 & 0.035{\scriptsize$\pm$0.010} \\*
PACT   & Greedy & \textbf{99.6}{\scriptsize$\pm$0.8} & \textbf{5.79}{\scriptsize$\pm$0.38} & \textbf{5.68}{\scriptsize$\pm$0.13} & 1.000 & 0.064{\scriptsize$\pm$0.053} \\*
\midrule
PARCO  & Sampling & \textbf{100.0}{\scriptsize$\pm$0.0} & 5.50{\scriptsize$\pm$0.06} & 5.50{\scriptsize$\pm$0.06} & 1.180 & 0.15{\scriptsize$\pm$0.06} \\*
MACSIM & Sampling & \textbf{100.0}{\scriptsize$\pm$0.0} & 6.32{\scriptsize$\pm$0.42} & 6.32{\scriptsize$\pm$0.42} & 1.356 & 0.68{\scriptsize$\pm$0.28} \\*
PACT   & Sampling & \textbf{100.0}{\scriptsize$\pm$0.0} & \textbf{4.66}{\scriptsize$\pm$0.05} & \textbf{4.66}{\scriptsize$\pm$0.05} & 1.000 & 5.45{\scriptsize$\pm$0.67} \\
\midrule
\multicolumn{7}{c}{\textbf{$128\times12\times24$} (ZST)}\\*
\noalign{\nobreak\hrule height\lightrulewidth\nobreak}
MAPPO  & Greedy & 95.7{\scriptsize$\pm$5.5} & 32.8{\scriptsize$\pm$16.4} & 20.6{\scriptsize$\pm$3.8} & 1.763 & 0.79{\scriptsize$\pm$0.66} \\*
IPPO   & Greedy & 97.2{\scriptsize$\pm$2.4} & 28.6{\scriptsize$\pm$10.3} & 21.3{\scriptsize$\pm$6.1} & 1.538 & 0.63{\scriptsize$\pm$0.48} \\*
VDN    & Greedy & \textbf{99.9}{\scriptsize$\pm$0.3} & 20.7{\scriptsize$\pm$4.5} & 20.3{\scriptsize$\pm$4.1} & 1.113 & 0.38{\scriptsize$\pm$0.15} \\*
DCG    & Greedy & 93.3{\scriptsize$\pm$17.8} & 70.0{\scriptsize$\pm$101.0} & 52.9{\scriptsize$\pm$69.2} & 3.763 & 0.99{\scriptsize$\pm$1.40} \\*
PARCO  & Greedy & 41.2{\scriptsize$\pm$24.2} & 195.1{\scriptsize$\pm$89.4} & 48.9{\scriptsize$\pm$36.0} & 10.489 & 0.11{\scriptsize$\pm$0.04} \\*
MACSIM & Greedy & 99.6{\scriptsize$\pm$0.7} & 34.8{\scriptsize$\pm$3.4} & 33.4{\scriptsize$\pm$3.0} & 1.871 & 0.047{\scriptsize$\pm$0.032} \\*
PACT   & Greedy & 99.7{\scriptsize$\pm$0.7} & \textbf{18.6}{\scriptsize$\pm$3.1} & \textbf{17.8}{\scriptsize$\pm$0.9} & 1.000 & 0.34{\scriptsize$\pm$0.18} \\*
\midrule
PARCO  & Sampling & 99.7{\scriptsize$\pm$0.7} & 17.8{\scriptsize$\pm$5.2} & 17.6{\scriptsize$\pm$4.6} & 1.148 & 2.67{\scriptsize$\pm$1.95} \\*
MACSIM & Sampling & \textbf{100.0}{\scriptsize$\pm$0.0} & 24.0{\scriptsize$\pm$2.9} & 24.0{\scriptsize$\pm$2.9} & 1.548 & 1.91{\scriptsize$\pm$1.63} \\*
PACT   & Sampling & \textbf{100.0}{\scriptsize$\pm$0.0} & \textbf{15.5}{\scriptsize$\pm$0.9} & \textbf{15.5}{\scriptsize$\pm$0.9} & 1.000 & 10.5{\scriptsize$\pm$1.6} \\
\midrule
\multicolumn{7}{c}{\textbf{$512\times24\times48$} (ZST)}\\*
\noalign{\nobreak\hrule height\lightrulewidth\nobreak}
MAPPO  & Greedy & OOM & OOM & OOM & OOM & OOM \\*
IPPO   & Greedy & OOM & OOM & OOM & OOM & OOM \\*
VDN    & Greedy & OOM & OOM & OOM & OOM & OOM \\*
DCG    & Greedy & OOM & OOM & OOM & OOM & OOM \\*
PARCO  & Greedy & 4.2{\scriptsize$\pm$8.4} & 1722.0{\scriptsize$\pm$555.4} & 225.5{\scriptsize$\pm$200.4} & 23.817 & 0.35{\scriptsize$\pm$0.06} \\*
MACSIM & Greedy & \textbf{100.0}{\scriptsize$\pm$0.0} & 90.2{\scriptsize$\pm$3.1} & 90.2{\scriptsize$\pm$3.1} & 1.248 & 0.18{\scriptsize$\pm$0.01} \\*
PACT   & Greedy & 99.0{\scriptsize$\pm$2.5} & \textbf{72.3}{\scriptsize$\pm$42.8} & \textbf{46.2}{\scriptsize$\pm$4.0} & 1.000 & 2.61{\scriptsize$\pm$2.71} \\*
\midrule
PARCO  & Sampling & 86.3{\scriptsize$\pm$10.3} & 372.6{\scriptsize$\pm$267.9} & 112.4{\scriptsize$\pm$116.5} & 7.408 & 9.83{\scriptsize$\pm$12.04} \\*
MACSIM & Sampling & \textbf{100.0}{\scriptsize$\pm$0.0} & 84.2{\scriptsize$\pm$13.3} & 84.2{\scriptsize$\pm$13.3} & 1.674 & 42.0{\scriptsize$\pm$8.9} \\*
PACT   & Sampling & \textbf{100.0}{\scriptsize$\pm$0.0} & \textbf{50.3}{\scriptsize$\pm$5.8} & \textbf{50.3}{\scriptsize$\pm$5.8} & 1.000 & 51.9{\scriptsize$\pm$14.6} \\
\midrule
\multicolumn{7}{c}{\textbf{$1024\times32\times64$} (ZST)}\\*
\noalign{\nobreak\hrule height\lightrulewidth\nobreak}
MAPPO  & Greedy & OOM & OOM & OOM & OOM & OOM \\*
IPPO   & Greedy & OOM & OOM & OOM & OOM & OOM \\*
VDN    & Greedy & OOM & OOM & OOM & OOM & OOM \\*
DCG    & Greedy & OOM & OOM & OOM & OOM & OOM \\*
PARCO  & Greedy & 4.5{\scriptsize$\pm$10.0} & 3999.2{\scriptsize$\pm$1871.4} & 472.0{\scriptsize$\pm$558.1} & 46.448 & 0.73{\scriptsize$\pm$0.15} \\*
MACSIM & Greedy & \textbf{100.0}{\scriptsize$\pm$0.0} & 131.3{\scriptsize$\pm$2.2} & 131.3{\scriptsize$\pm$2.2} & 1.525 & 2.20{\scriptsize$\pm$0.15} \\*
PACT   & Greedy & 99.6{\scriptsize$\pm$1.3} & \textbf{86.1}{\scriptsize$\pm$29.2} & \textbf{70.5}{\scriptsize$\pm$11.1} & 1.000 & 11.0{\scriptsize$\pm$14.7} \\*
\midrule
PARCO  & Sampling & 71.3{\scriptsize$\pm$16.2} & 1341.6{\scriptsize$\pm$349.6} & 187.1{\scriptsize$\pm$214.0} & 15.971 & 27.5{\scriptsize$\pm$27.9} \\*
MACSIM & Sampling & \textbf{100.0}{\scriptsize$\pm$0.0} & 125.8{\scriptsize$\pm$15.0} & 125.8{\scriptsize$\pm$15.0} & 1.498 & 381.7{\scriptsize$\pm$31.0} \\*
PACT   & Sampling & \textbf{100.0}{\scriptsize$\pm$0.0} & \textbf{84.0}{\scriptsize$\pm$10.8} & \textbf{84.0}{\scriptsize$\pm$10.8} & 1.000 & 136.9{\scriptsize$\pm$27.9} \\
\end{longtable}
}

{\small
\setlength{\aboverulesep}{0pt}
\setlength{\belowrulesep}{0pt}
\renewcommand{\arraystretch}{1.0}
\setlength{\tabcolsep}{3pt}
\setlength{\completeTableColumnWidth}{\dimexpr0.98\textwidth-12\tabcolsep-6\arrayrulewidth\relax}
\begin{longtable}{@{}>{\raggedright\arraybackslash}m{0.11\completeTableColumnWidth}|>{\raggedright\arraybackslash}m{0.12\completeTableColumnWidth}|>{\centering\arraybackslash}m{0.14\completeTableColumnWidth}|>{\centering\arraybackslash}m{0.17\completeTableColumnWidth}|>{\centering\arraybackslash}m{0.17\completeTableColumnWidth}|>{\centering\arraybackslash}m{0.12\completeTableColumnWidth}|>{\centering\arraybackslash}m{0.17\completeTableColumnWidth}@{}}
\caption{Complete results for learning-based methods on DFJSP. The
first configuration is in-distribution (ID), and the remaining configurations are evaluated
zero-shot (ZST).}
\label{tab:complete-dfjsp}\\
\toprule
Method & Evaluation & Success (\%) $\uparrow$ & Cost $\downarrow$ & Objective $\downarrow$ & Perf. ratio $\downarrow$ & Time (s) $\downarrow$ \\
\midrule
\endfirsthead
\multicolumn{7}{c}{\tablename~\thetable\ continued}\\
\toprule
Method & Evaluation & Success (\%) $\uparrow$ & Cost $\downarrow$ & Objective $\downarrow$ & Perf. ratio $\downarrow$ & Time (s) $\downarrow$ \\
\midrule
\endhead
\bottomrule
\endfoot
\multicolumn{7}{c}{\textbf{$20\times8\times6$} (ID)}\\*
\noalign{\nobreak\hrule height\lightrulewidth\nobreak}
MAPPO  & Greedy & \textbf{100.0}{\scriptsize$\pm$0.0} & 170.2{\scriptsize$\pm$3.2} & 170.2{\scriptsize$\pm$3.2} & 1.047 & 0.022{\scriptsize$\pm$0.008} \\*
IPPO   & Greedy & 96.1{\scriptsize$\pm$9.4} & 207.7{\scriptsize$\pm$83.7} & 176.2{\scriptsize$\pm$7.1} & 1.278 & 0.051{\scriptsize$\pm$0.017} \\*
VDN    & Greedy & 54.4{\scriptsize$\pm$48.2} & 632.1{\scriptsize$\pm$486.7} & 181.7{\scriptsize$\pm$6.4} & 3.890 & 0.056{\scriptsize$\pm$0.028} \\*
DCG    & Greedy & 15.1{\scriptsize$\pm$32.1} & 986.6{\scriptsize$\pm$411.5} & 187.8{\scriptsize$\pm$14.6} & 6.071 & 0.088{\scriptsize$\pm$0.004} \\*
PARCO  & Greedy & \textbf{100.0}{\scriptsize$\pm$0.0} & 170.5{\scriptsize$\pm$0.4} & 170.5{\scriptsize$\pm$0.4} & 1.049 & 0.004{\scriptsize$\pm$0.001} \\*
MACSIM & Greedy & 54.7{\scriptsize$\pm$20.3} & 608.3{\scriptsize$\pm$212.0} & 166.1{\scriptsize$\pm$4.3} & 3.743 & 0.18{\scriptsize$\pm$0.03} \\*
PACT   & Greedy & \textbf{100.0}{\scriptsize$\pm$0.0} & \textbf{162.5}{\scriptsize$\pm$2.9} & \textbf{162.5}{\scriptsize$\pm$2.9} & 1.000 & 0.093{\scriptsize$\pm$0.042} \\*
\midrule
PARCO  & Sampling & \textbf{100.0}{\scriptsize$\pm$0.0} & 165.0{\scriptsize$\pm$0.3} & 165.0{\scriptsize$\pm$0.3} & 1.058 & 0.087{\scriptsize$\pm$0.001} \\*
MACSIM & Sampling & \textbf{100.0}{\scriptsize$\pm$0.0} & 156.2{\scriptsize$\pm$1.1} & 156.2{\scriptsize$\pm$1.1} & 1.002 & 0.20{\scriptsize$\pm$0.09} \\*
PACT   & Sampling & \textbf{100.0}{\scriptsize$\pm$0.0} & \textbf{155.9}{\scriptsize$\pm$1.4} & \textbf{155.9}{\scriptsize$\pm$1.4} & 1.000 & 10.5{\scriptsize$\pm$3.7} \\
\midrule
\multicolumn{7}{c}{\textbf{$30\times14\times6$} (ZST)}\\*
\noalign{\nobreak\hrule height\lightrulewidth\nobreak}
MAPPO  & Greedy & \textbf{100.0}{\scriptsize$\pm$0.0} & 148.4{\scriptsize$\pm$3.2} & 148.4{\scriptsize$\pm$3.2} & 1.049 & 0.021{\scriptsize$\pm$0.009} \\*
IPPO   & Greedy & 99.7{\scriptsize$\pm$0.7} & 156.1{\scriptsize$\pm$12.5} & 152.4{\scriptsize$\pm$4.8} & 1.103 & 0.054{\scriptsize$\pm$0.017} \\*
VDN    & Greedy & 66.5{\scriptsize$\pm$46.3} & 688.7{\scriptsize$\pm$748.7} & 158.3{\scriptsize$\pm$5.6} & 4.867 & 0.076{\scriptsize$\pm$0.049} \\*
DCG    & Greedy & 13.4{\scriptsize$\pm$28.1} & 1476.8{\scriptsize$\pm$604.8} & 217.6{\scriptsize$\pm$76.6} & 10.437 & 0.13{\scriptsize$\pm$$<$0.01} \\*
PARCO  & Greedy & \textbf{100.0}{\scriptsize$\pm$0.0} & 146.0{\scriptsize$\pm$0.5} & 146.0{\scriptsize$\pm$0.5} & 1.032 & 0.004{\scriptsize$\pm$0.001} \\*
MACSIM & Greedy & 70.1{\scriptsize$\pm$21.7} & 596.9{\scriptsize$\pm$311.7} & \textbf{141.5}{\scriptsize$\pm$4.3} & 4.218 & 0.21{\scriptsize$\pm$0.05} \\*
PACT   & Greedy & \textbf{100.0}{\scriptsize$\pm$0.0} & \textbf{141.5}{\scriptsize$\pm$3.6} & \textbf{141.5}{\scriptsize$\pm$3.6} & 1.000 & 0.10{\scriptsize$\pm$0.05} \\*
\midrule
PARCO  & Sampling & \textbf{100.0}{\scriptsize$\pm$0.0} & 142.8{\scriptsize$\pm$0.5} & 142.8{\scriptsize$\pm$0.5} & 1.065 & 0.096{\scriptsize$\pm$0.001} \\*
MACSIM & Sampling & \textbf{100.0}{\scriptsize$\pm$0.0} & 136.4{\scriptsize$\pm$1.2} & 136.4{\scriptsize$\pm$1.2} & 1.017 & 0.25{\scriptsize$\pm$0.09} \\*
PACT   & Sampling & \textbf{100.0}{\scriptsize$\pm$0.0} & \textbf{134.1}{\scriptsize$\pm$1.9} & \textbf{134.1}{\scriptsize$\pm$1.9} & 1.000 & 10.5{\scriptsize$\pm$4.3} \\
\midrule
\multicolumn{7}{c}{\textbf{$50\times16\times8$} (ZST)}\\*
\noalign{\nobreak\hrule height\lightrulewidth\nobreak}
MAPPO  & Greedy & \textbf{100.0}{\scriptsize$\pm$0.0} & 245.1{\scriptsize$\pm$5.9} & 245.1{\scriptsize$\pm$5.9} & 1.051 & 0.070{\scriptsize$\pm$0.033} \\*
IPPO   & Greedy & 98.7{\scriptsize$\pm$3.0} & 295.6{\scriptsize$\pm$86.2} & 261.6{\scriptsize$\pm$11.0} & 1.268 & 0.24{\scriptsize$\pm$0.11} \\*
VDN    & Greedy & 57.7{\scriptsize$\pm$50.0} & 1641.3{\scriptsize$\pm$1647.5} & 256.2{\scriptsize$\pm$18.6} & 7.038 & 0.33{\scriptsize$\pm$0.27} \\*
DCG    & Greedy & 7.4{\scriptsize$\pm$15.9} & 3089.9{\scriptsize$\pm$1041.2} & 335.3{\scriptsize$\pm$71.8} & 13.250 & 0.61{\scriptsize$\pm$0.01} \\*
PARCO  & Greedy & 99.7{\scriptsize$\pm$0.9} & 258.4{\scriptsize$\pm$2.7} & 253.7{\scriptsize$\pm$2.0} & 1.108 & 0.015{\scriptsize$\pm$0.024} \\*
MACSIM & Greedy & 60.2{\scriptsize$\pm$29.8} & 1241.3{\scriptsize$\pm$195.5} & 247.3{\scriptsize$\pm$9.2} & 5.323 & 0.48{\scriptsize$\pm$0.13} \\*
PACT   & Greedy & \textbf{100.0}{\scriptsize$\pm$0.0} & \textbf{233.2}{\scriptsize$\pm$3.6} & \textbf{233.2}{\scriptsize$\pm$3.6} & 1.000 & 0.25{\scriptsize$\pm$0.15} \\*
\midrule
PARCO  & Sampling & \textbf{100.0}{\scriptsize$\pm$0.0} & 249.7{\scriptsize$\pm$1.7} & 249.7{\scriptsize$\pm$1.7} & 1.074 & 0.18{\scriptsize$\pm$0.01} \\*
MACSIM & Sampling & \textbf{100.0}{\scriptsize$\pm$0.0} & 242.3{\scriptsize$\pm$1.8} & 242.3{\scriptsize$\pm$1.8} & 1.043 & 0.93{\scriptsize$\pm$0.93} \\*
PACT   & Sampling & \textbf{100.0}{\scriptsize$\pm$0.0} & \textbf{232.4}{\scriptsize$\pm$2.6} & \textbf{232.4}{\scriptsize$\pm$2.6} & 1.000 & 23.6{\scriptsize$\pm$11.0} \\
\midrule
\multicolumn{7}{c}{\textbf{$80\times20\times8$} (ZST)}\\*
\noalign{\nobreak\hrule height\lightrulewidth\nobreak}
MAPPO  & Greedy & \textbf{100.0}{\scriptsize$\pm$0.0} & 299.0{\scriptsize$\pm$9.8} & 299.0{\scriptsize$\pm$9.8} & 1.053 & 0.24{\scriptsize$\pm$0.12} \\*
IPPO   & Greedy & 98.6{\scriptsize$\pm$3.5} & 380.1{\scriptsize$\pm$162.5} & 319.9{\scriptsize$\pm$16.3} & 1.339 & 0.87{\scriptsize$\pm$0.49} \\*
VDN    & Greedy & 59.6{\scriptsize$\pm$51.3} & 2498.4{\scriptsize$\pm$2780.2} & 318.3{\scriptsize$\pm$18.6} & 8.800 & 1.27{\scriptsize$\pm$1.12} \\*
DCG    & Greedy & 5.4{\scriptsize$\pm$11.4} & 5017.7{\scriptsize$\pm$1604.1} & 452.9{\scriptsize$\pm$148.6} & 17.674 & 2.72{\scriptsize$\pm$0.02} \\*
PARCO  & Greedy & 97.4{\scriptsize$\pm$5.9} & 336.2{\scriptsize$\pm$32.6} & 314.6{\scriptsize$\pm$4.6} & 1.184 & 0.054{\scriptsize$\pm$0.084} \\*
MACSIM & Greedy & 58.0{\scriptsize$\pm$32.2} & 1305.2{\scriptsize$\pm$1130.6} & 312.5{\scriptsize$\pm$11.6} & 4.597 & 0.84{\scriptsize$\pm$0.24} \\*
PACT   & Greedy & \textbf{100.0}{\scriptsize$\pm$0.0} & \textbf{283.9}{\scriptsize$\pm$4.7} & \textbf{283.9}{\scriptsize$\pm$4.7} & 1.000 & 0.77{\scriptsize$\pm$0.50} \\*
\midrule
PARCO  & Sampling & \textbf{100.0}{\scriptsize$\pm$0.0} & 314.5{\scriptsize$\pm$3.9} & 314.5{\scriptsize$\pm$3.9} & 1.125 & 0.28{\scriptsize$\pm$0.03} \\*
MACSIM & Sampling & \textbf{100.0}{\scriptsize$\pm$0.0} & 307.2{\scriptsize$\pm$2.3} & 307.2{\scriptsize$\pm$2.3} & 1.099 & 1.83{\scriptsize$\pm$1.75} \\*
PACT   & Sampling & \textbf{100.0}{\scriptsize$\pm$0.0} & \textbf{279.6}{\scriptsize$\pm$3.5} & \textbf{279.6}{\scriptsize$\pm$3.5} & 1.000 & 54.4{\scriptsize$\pm$28.7} \\
\midrule
\multicolumn{7}{c}{\textbf{$120\times24\times10$} (ZST)}\\*
\noalign{\nobreak\hrule height\lightrulewidth\nobreak}
MAPPO  & Greedy & \textbf{100.0}{\scriptsize$\pm$0.0} & 429.2{\scriptsize$\pm$20.5} & 429.2{\scriptsize$\pm$20.5} & 1.077 & 0.84{\scriptsize$\pm$0.45} \\*
IPPO   & Greedy & 98.5{\scriptsize$\pm$4.4} & 584.4{\scriptsize$\pm$354.4} & 469.7{\scriptsize$\pm$30.9} & 1.467 & 3.45{\scriptsize$\pm$2.43} \\*
VDN    & Greedy & 59.9{\scriptsize$\pm$51.6} & 4311.0{\scriptsize$\pm$4961.6} & 467.5{\scriptsize$\pm$27.1} & 10.821 & 5.54{\scriptsize$\pm$5.48} \\*
DCG    & Greedy & 0.5{\scriptsize$\pm$1.6} & 9296.8{\scriptsize$\pm$2400.1} & 1030.2{\scriptsize$\pm$1062.3} & 23.335 & 13.1{\scriptsize$\pm$0.1} \\*
PARCO  & Greedy & 7.6{\scriptsize$\pm$13.7} & 2578.1{\scriptsize$\pm$608.2} & 439.1{\scriptsize$\pm$23.1} & 6.471 & 0.35{\scriptsize$\pm$0.08} \\*
MACSIM & Greedy & 61.2{\scriptsize$\pm$32.3} & 1866.3{\scriptsize$\pm$641.7} & 462.0{\scriptsize$\pm$19.7} & 4.684 & 1.64{\scriptsize$\pm$0.52} \\*
PACT   & Greedy & \textbf{100.0}{\scriptsize$\pm$0.0} & \textbf{398.4}{\scriptsize$\pm$6.8} & \textbf{398.4}{\scriptsize$\pm$6.8} & 1.000 & 2.41{\scriptsize$\pm$1.56} \\*
\midrule
PARCO  & Sampling & \textbf{100.0}{\scriptsize$\pm$0.0} & 490.3{\scriptsize$\pm$13.6} & 490.3{\scriptsize$\pm$13.6} & 1.237 & 2.30{\scriptsize$\pm$2.86} \\*
MACSIM & Sampling & \textbf{100.0}{\scriptsize$\pm$0.0} & 458.3{\scriptsize$\pm$2.9} & 458.3{\scriptsize$\pm$2.9} & 1.156 & 5.24{\scriptsize$\pm$9.66} \\*
PACT   & Sampling & \textbf{100.0}{\scriptsize$\pm$0.0} & \textbf{396.5}{\scriptsize$\pm$4.9} & \textbf{396.5}{\scriptsize$\pm$4.9} & 1.000 & 112.1{\scriptsize$\pm$27.3} \\
\end{longtable}
}

{\small
\setlength{\aboverulesep}{0pt}
\setlength{\belowrulesep}{0pt}
\renewcommand{\arraystretch}{1.0}
\setlength{\tabcolsep}{3pt}
\setlength{\completeTableColumnWidth}{\dimexpr0.98\textwidth-12\tabcolsep-6\arrayrulewidth\relax}
\begin{longtable}{@{}>{\raggedright\arraybackslash}m{0.11\completeTableColumnWidth}|>{\raggedright\arraybackslash}m{0.12\completeTableColumnWidth}|>{\centering\arraybackslash}m{0.14\completeTableColumnWidth}|>{\centering\arraybackslash}m{0.17\completeTableColumnWidth}|>{\centering\arraybackslash}m{0.17\completeTableColumnWidth}|>{\centering\arraybackslash}m{0.12\completeTableColumnWidth}|>{\centering\arraybackslash}m{0.17\completeTableColumnWidth}@{}}
\caption{Complete results for learning-based methods on NCFFSP. The
first configuration is in-distribution (ID), and the remaining configurations are evaluated
zero-shot (ZST).}
\label{tab:complete-ncffsp}\\
\toprule
Method & Evaluation & Success (\%) $\uparrow$ & Cost $\downarrow$ & Objective $\downarrow$ & Perf. ratio $\downarrow$ & Time (s) $\downarrow$ \\
\midrule
\endfirsthead
\multicolumn{7}{c}{\tablename~\thetable\ continued}\\
\toprule
Method & Evaluation & Success (\%) $\uparrow$ & Cost $\downarrow$ & Objective $\downarrow$ & Perf. ratio $\downarrow$ & Time (s) $\downarrow$ \\
\midrule
\endhead
\bottomrule
\endfoot
\multicolumn{7}{c}{\textbf{$20\times4\times3$} (ID)}\\*
\noalign{\nobreak\hrule height\lightrulewidth\nobreak}
MAPPO  & Greedy & 97.4{\scriptsize$\pm$3.5} & 46.5{\scriptsize$\pm$2.3} & 44.2{\scriptsize$\pm$1.0} & 1.120 & 0.019{\scriptsize$\pm$0.006} \\*
IPPO   & Greedy & 98.5{\scriptsize$\pm$1.9} & 47.3{\scriptsize$\pm$1.7} & 45.4{\scriptsize$\pm$1.5} & 1.140 & 0.017{\scriptsize$\pm$0.004} \\*
VDN    & Greedy & 97.8{\scriptsize$\pm$2.2} & 50.0{\scriptsize$\pm$2.1} & 48.6{\scriptsize$\pm$1.5} & 1.205 & 0.017{\scriptsize$\pm$0.005} \\*
DCG    & Greedy & 71.7{\scriptsize$\pm$31.6} & 84.1{\scriptsize$\pm$50.6} & 56.1{\scriptsize$\pm$17.0} & 2.027 & 0.036{\scriptsize$\pm$0.011} \\*
PARCO  & Greedy & 99.3{\scriptsize$\pm$0.7} & 42.9{\scriptsize$\pm$0.5} & 42.5{\scriptsize$\pm$0.4} & 1.034 & 0.005{\scriptsize$\pm$0.002} \\*
MACSIM & Greedy & 82.9{\scriptsize$\pm$32.5} & 64.7{\scriptsize$\pm$43.0} & 45.9{\scriptsize$\pm$5.9} & 1.559 & 0.067{\scriptsize$\pm$0.022} \\*
PACT   & Greedy & \textbf{100.0}{\scriptsize$\pm$0.0} & \textbf{41.5}{\scriptsize$\pm$0.7} & \textbf{41.5}{\scriptsize$\pm$0.7} & 1.000 & 0.049{\scriptsize$\pm$0.019} \\*
\midrule
PARCO  & Sampling & \textbf{100.0}{\scriptsize$\pm$0.0} & 37.3{\scriptsize$\pm$0.2} & 37.3{\scriptsize$\pm$0.2} & 1.030 & 0.085{\scriptsize$\pm$0.002} \\*
MACSIM & Sampling & \textbf{100.0}{\scriptsize$\pm$0.0} & 39.0{\scriptsize$\pm$1.5} & 39.0{\scriptsize$\pm$1.5} & 1.077 & 0.21{\scriptsize$\pm$0.03} \\*
PACT   & Sampling & \textbf{100.0}{\scriptsize$\pm$0.0} & \textbf{36.2}{\scriptsize$\pm$0.6} & \textbf{36.2}{\scriptsize$\pm$0.6} & 1.000 & 5.35{\scriptsize$\pm$0.26} \\
\midrule
\multicolumn{7}{c}{\textbf{$30\times6\times3$} (ZST)}\\*
\noalign{\nobreak\hrule height\lightrulewidth\nobreak}
MAPPO  & Greedy & 99.2{\scriptsize$\pm$1.3} & 38.0{\scriptsize$\pm$1.9} & 37.3{\scriptsize$\pm$0.8} & 1.058 & 0.020{\scriptsize$\pm$0.007} \\*
IPPO   & Greedy & 99.3{\scriptsize$\pm$1.3} & 39.3{\scriptsize$\pm$1.0} & 38.5{\scriptsize$\pm$1.2} & 1.095 & 0.020{\scriptsize$\pm$0.007} \\*
VDN    & Greedy & 96.8{\scriptsize$\pm$2.8} & 44.2{\scriptsize$\pm$3.1} & 41.3{\scriptsize$\pm$1.4} & 1.231 & 0.027{\scriptsize$\pm$0.010} \\*
DCG    & Greedy & 58.9{\scriptsize$\pm$25.5} & 102.4{\scriptsize$\pm$53.7} & 52.3{\scriptsize$\pm$13.4} & 2.852 & 0.069{\scriptsize$\pm$0.005} \\*
PARCO  & Greedy & 98.8{\scriptsize$\pm$0.9} & 49.8{\scriptsize$\pm$1.2} & 48.7{\scriptsize$\pm$1.4} & 1.387 & 0.009{\scriptsize$\pm$0.005} \\*
MACSIM & Greedy & 91.7{\scriptsize$\pm$23.3} & 55.4{\scriptsize$\pm$46.4} & 41.2{\scriptsize$\pm$5.7} & 1.543 & 0.057{\scriptsize$\pm$0.032} \\*
PACT   & Greedy & \textbf{100.0}{\scriptsize$\pm$0.0} & \textbf{35.9}{\scriptsize$\pm$1.9} & \textbf{35.9}{\scriptsize$\pm$1.9} & 1.000 & 0.063{\scriptsize$\pm$0.041} \\*
\midrule
PARCO  & Sampling & \textbf{100.0}{\scriptsize$\pm$0.0} & 39.6{\scriptsize$\pm$0.4} & 39.6{\scriptsize$\pm$0.4} & 1.119 & 0.099{\scriptsize$\pm$0.002} \\*
MACSIM & Sampling & \textbf{100.0}{\scriptsize$\pm$0.0} & 36.1{\scriptsize$\pm$2.3} & 36.1{\scriptsize$\pm$2.3} & 1.020 & 0.32{\scriptsize$\pm$0.04} \\*
PACT   & Sampling & \textbf{100.0}{\scriptsize$\pm$0.0} & \textbf{35.4}{\scriptsize$\pm$0.9} & \textbf{35.4}{\scriptsize$\pm$0.9} & 1.000 & 6.17{\scriptsize$\pm$0.29} \\
\midrule
\multicolumn{7}{c}{\textbf{$50\times7\times3$} (ZST)}\\*
\noalign{\nobreak\hrule height\lightrulewidth\nobreak}
MAPPO  & Greedy & 99.3{\scriptsize$\pm$1.3} & 49.4{\scriptsize$\pm$2.8} & \textbf{48.0}{\scriptsize$\pm$0.7} & 1.012 & 0.065{\scriptsize$\pm$0.027} \\*
IPPO   & Greedy & 98.9{\scriptsize$\pm$1.9} & 52.6{\scriptsize$\pm$2.5} & 49.7{\scriptsize$\pm$2.6} & 1.078 & 0.070{\scriptsize$\pm$0.032} \\*
VDN    & Greedy & 93.2{\scriptsize$\pm$8.3} & 63.9{\scriptsize$\pm$12.5} & 53.5{\scriptsize$\pm$2.4} & 1.309 & 0.12{\scriptsize$\pm$0.07} \\*
DCG    & Greedy & 59.0{\scriptsize$\pm$30.0} & 152.4{\scriptsize$\pm$79.0} & 75.4{\scriptsize$\pm$30.9} & 3.123 & 0.26{\scriptsize$\pm$0.03} \\*
PARCO  & Greedy & 98.8{\scriptsize$\pm$2.2} & 66.8{\scriptsize$\pm$3.5} & 65.3{\scriptsize$\pm$1.3} & 1.369 & 0.010{\scriptsize$\pm$0.006} \\*
MACSIM & Greedy & 86.6{\scriptsize$\pm$30.5} & 98.4{\scriptsize$\pm$104.2} & 61.3{\scriptsize$\pm$19.1} & 2.016 & 0.11{\scriptsize$\pm$0.07} \\*
PACT   & Greedy & \textbf{99.9}{\scriptsize$\pm$0.3} & \textbf{48.8}{\scriptsize$\pm$2.4} & 48.4{\scriptsize$\pm$1.1} & 1.000 & 0.17{\scriptsize$\pm$0.14} \\*
\midrule
PARCO  & Sampling & \textbf{100.0}{\scriptsize$\pm$0.0} & 57.4{\scriptsize$\pm$2.3} & 57.4{\scriptsize$\pm$2.3} & 1.281 & 0.13{\scriptsize$\pm$0.09} \\*
MACSIM & Sampling & \textbf{100.0}{\scriptsize$\pm$0.0} & 50.7{\scriptsize$\pm$4.0} & 50.7{\scriptsize$\pm$4.0} & 1.132 & 0.71{\scriptsize$\pm$0.09} \\*
PACT   & Sampling & \textbf{100.0}{\scriptsize$\pm$0.0} & \textbf{44.8}{\scriptsize$\pm$1.5} & \textbf{44.8}{\scriptsize$\pm$1.5} & 1.000 & 11.7{\scriptsize$\pm$0.6} \\
\midrule
\multicolumn{7}{c}{\textbf{$80\times6\times4$} (ZST)}\\*
\noalign{\nobreak\hrule height\lightrulewidth\nobreak}
MAPPO  & Greedy & 98.2{\scriptsize$\pm$3.3} & 94.5{\scriptsize$\pm$11.8} & 88.5{\scriptsize$\pm$1.3} & 1.133 & 0.33{\scriptsize$\pm$0.24} \\*
IPPO   & Greedy & 97.8{\scriptsize$\pm$4.5} & 99.8{\scriptsize$\pm$14.1} & 91.6{\scriptsize$\pm$5.0} & 1.197 & 0.30{\scriptsize$\pm$0.24} \\*
VDN    & Greedy & 84.4{\scriptsize$\pm$11.7} & 151.0{\scriptsize$\pm$42.7} & 98.8{\scriptsize$\pm$2.9} & 1.811 & 1.04{\scriptsize$\pm$0.42} \\*
DCG    & Greedy & 50.3{\scriptsize$\pm$18.1} & 376.7{\scriptsize$\pm$137.6} & 141.8{\scriptsize$\pm$53.5} & 4.517 & 1.64{\scriptsize$\pm$0.03} \\*
PARCO  & Greedy & N/A & N/A & N/A & N/A & N/A \\*
MACSIM & Greedy & 82.7{\scriptsize$\pm$28.5} & 212.6{\scriptsize$\pm$191.5} & 114.4{\scriptsize$\pm$27.9} & 2.549 & 0.29{\scriptsize$\pm$0.16} \\*
PACT   & Greedy & \textbf{99.7}{\scriptsize$\pm$0.7} & \textbf{83.4}{\scriptsize$\pm$8.9} & \textbf{81.1}{\scriptsize$\pm$2.3} & 1.000 & 0.51{\scriptsize$\pm$0.57} \\*
\midrule
PARCO  & Sampling & N/A & N/A & N/A & N/A & N/A \\*
MACSIM & Sampling & \textbf{100.0}{\scriptsize$\pm$0.0} & 97.3{\scriptsize$\pm$7.1} & 97.3{\scriptsize$\pm$7.1} & 1.179 & 2.09{\scriptsize$\pm$0.66} \\*
PACT   & Sampling & \textbf{100.0}{\scriptsize$\pm$0.0} & \textbf{82.5}{\scriptsize$\pm$2.9} & \textbf{82.5}{\scriptsize$\pm$2.9} & 1.000 & 32.5{\scriptsize$\pm$1.6} \\
\midrule
\multicolumn{7}{c}{\textbf{$120\times8\times4$} (ZST)}\\*
\noalign{\nobreak\hrule height\lightrulewidth\nobreak}
MAPPO  & Greedy & 98.4{\scriptsize$\pm$3.5} & 93.8{\scriptsize$\pm$16.1} & 86.2{\scriptsize$\pm$2.5} & 1.082 & 1.13{\scriptsize$\pm$1.22} \\*
IPPO   & Greedy & 98.6{\scriptsize$\pm$3.5} & 97.1{\scriptsize$\pm$16.2} & 90.2{\scriptsize$\pm$4.4} & 1.120 & 1.05{\scriptsize$\pm$1.15} \\*
VDN    & Greedy & 81.6{\scriptsize$\pm$13.2} & 186.1{\scriptsize$\pm$65.9} & 96.8{\scriptsize$\pm$2.8} & 2.146 & 4.62{\scriptsize$\pm$1.60} \\*
DCG    & Greedy & 48.4{\scriptsize$\pm$27.9} & 513.9{\scriptsize$\pm$241.2} & 186.0{\scriptsize$\pm$99.1} & 5.927 & 6.98{\scriptsize$\pm$0.35} \\*
PARCO  & Greedy & N/A & N/A & N/A & N/A & N/A \\*
MACSIM & Greedy & 89.1{\scriptsize$\pm$23.1} & 213.9{\scriptsize$\pm$222.1} & 123.6{\scriptsize$\pm$29.2} & 2.467 & 0.35{\scriptsize$\pm$0.29} \\*
PACT   & Greedy & \textbf{99.4}{\scriptsize$\pm$1.1} & \textbf{86.7}{\scriptsize$\pm$22.6} & \textbf{80.4}{\scriptsize$\pm$2.4} & 1.000 & 1.84{\scriptsize$\pm$2.40} \\*
\midrule
PARCO  & Sampling & N/A & N/A & N/A & N/A & N/A \\*
MACSIM & Sampling & \textbf{100.0}{\scriptsize$\pm$0.0} & 109.6{\scriptsize$\pm$8.9} & 109.6{\scriptsize$\pm$8.9} & 1.280 & 3.87{\scriptsize$\pm$0.62} \\*
PACT   & Sampling & \textbf{100.0}{\scriptsize$\pm$0.0} & \textbf{85.6}{\scriptsize$\pm$4.4} & \textbf{85.6}{\scriptsize$\pm$4.4} & 1.000 & 97.0{\scriptsize$\pm$10.0} \\
\end{longtable}
}

\subsection{Runtime Analysis}
\label{app:runtime-analysis}

Table~\ref{tab:runtime-breakdown} decomposes PACT's end-to-end runtime into environment stepping, policy inference, and other costs. 
Policy inference is further divided into encoding, candidate plan generation, trace-graph coordination, plan selection, and remaining policy operations. 
We report the ID and largest ZST configuration for each task to show how the distribution changes with problem scale.

The runtime profiles reflect both problem scale and the planning horizon $H_p$. 
On MACTP, where $H_p=4$, plan generation accounts for 38.02\% on the ID configuration and remains at 42.15\% on the largest ZST configuration because each of the $K$ candidates requires a multi-step autoregressive virtual rollout.
By contrast, DFJSP and NCFFSP use $H_p=1$, so plan generation ends after the initial candidate actions and its share falls from 23.82\% and 27.49\% to 3.74\% and 5.27\% as the instances grow.
Their larger graph states instead make encoding dominant, accounting for 86.37\% and 87.76\% of runtime. 
Encoding also increases from 11.91\% to 43.77\% on MACTP, but its longer rollouts keep plan generation comparably expensive. 
The relative runtime of coordination decreases from 9.95-13.25\% on the ID configurations to 2.81-4.42\% on the largest configurations, while selection remains below 1.3\% across the evaluated configurations.

\begin{table}[H]
\centering
\caption{Breakdown of PACT's end-to-end runtime. Component columns report percentages of the end-to-end runtime, and the final column reports time per instance.}
\label{tab:runtime-breakdown}
\setlength{\aboverulesep}{0pt}
\setlength{\belowrulesep}{0pt}
\renewcommand{\arraystretch}{1}
\begin{adjustbox}{max width=\textwidth,center}
\begin{tabular}{@{}l|l|r|rrrrr|r|r@{}}
\toprule
\multicolumn{1}{c|}{\multirow{2}{*}{Task}} &
\multicolumn{1}{c|}{\multirow{2}{*}{Configuration}} &
\multicolumn{1}{c|}{\multirow{2}{*}{{Env. step}}} &
\multicolumn{5}{c|}{{Policy}} &
\multicolumn{1}{c|}{\multirow{2}{*}{{Other}}} &
\multirow{2}{*}{Time (s)} \\
\cmidrule{4-8}
 & & & {Encoder} & {Plan} & {Coord.} & {Selection} & {Other} & & \\
\midrule
\multirow{2}{*}{MACTP} & $16\times4\times6$ (ID)       & {15.77\%} & {11.91\%} & {38.02\%} & {9.95\%} & {0.5\%} & {15.45\%} & {8.39\%} & {0.056} \\*
\cmidrule{2-10}
 & $1024\times32\times64$ (ZST)  & {5.32\%} & {43.77\%} & {42.15\%} & {2.81\%} & {0.14\%} & {4.5\%} & {1.31\%} & {11.0} \\*
\midrule
\multirow{2}{*}{DFJSP} & $20\times8\times6$ (ID)       & {19.04\%} & {15.35\%} & {23.82\%} & {12.92\%} & {1.28\%} & {23.5\%} & {4.08\%} & {0.093} \\*
\cmidrule{2-10}
 & $120\times24\times10$ (ZST)   & {2.38\%} & {86.37\%} & {3.74\%} & {4.42\%} & {0.07\%} & {2.51\%} & {0.53\%} & {2.41} \\*
\midrule
\multirow{2}{*}{NCFFSP} & $20\times4\times3$ (ID)       & {19.71\%} & {15.84\%} & {27.49\%} & {13.25\%} & {0.69\%} & {18.59\%} & {4.43\%} & {0.049} \\*
\cmidrule{2-10}
 & $120\times8\times4$ (ZST)     & {2.17\%} & {87.76\%} & {5.27\%} & {3.01\%} & {0.02\%} & {1.26\%} & {0.52\%} & {1.84} \\*
\bottomrule
\end{tabular}
\end{adjustbox}
\end{table}

\FloatBarrier

\subsection{Ablation Study Details}
\label{app:ablation}

In this section, we provide further details regarding the evaluation protocol and results of the ablation study reported in Section~\ref{sec:ablation} in the main text.

All ablation experiments are conducted with greedy evaluation on the largest zero-shot transfer configuration of each problem. We use the PACT model with $K=4$ candidate plans and $R=2$ coordination rounds as the reference model. 
We define the performance ratio as cost divided by the reference model's mean cost on the same problem, and compute the time ratio in the same way using end-to-end runtime. 
Both ratios equal one for the reference model, and lower values are better. 
We present the ablation results in Figure~\ref{fig:ablation}. For each method or configuration, bars show mean performance ratios across seeds, with error bars indicating one standard deviation, while lines show mean time ratios.

\noindent \textbf{Effect of Coordination.} We compare PACT, PACT without coordination, and MAPPO in Figure~\ref{fig:ablation}(\subref{fig:ablation-graph}). PACT without coordination omits message passing on the plan-augmented trace-graph, so each agent selects its candidate plan independently. PACT achieves the lowest mean performance ratio on all three problems. Removing coordination substantially degrades performance, particularly on MACTP and NCFFSP, and makes it worse than MAPPO on DFJSP and NCFFSP. This indicates that the gains arise from using shared elements to mediate candidate selection rather than from candidate generation alone.

\noindent \textbf{Number of Candidate Plans.} We vary the number of candidate plans $K\in\{1,4,8,16\}$ in Figure~\ref{fig:ablation}(\subref{fig:ablation-k}). With $K=1$, each agent exposes only one plan to the trace-graph, so coordination cannot redirect it to an alternative. This setting degrades performance on every problem and produces high variability on NCFFSP, whereas four candidates provide most of the observed improvement. Larger sets yield no consistent gain beyond the variation across seeds. The time ratio reflects both per-step computation and episode length, so longer episodes can offset the savings from fewer candidates. The sharp escalation in the time ratio on MACTP at large $K$ is consistent with $H_p=4$, compared with $H_p=1$ on DFJSP and NCFFSP.

\noindent \textbf{Number of Coordination Rounds.} We vary the number of coordination rounds $R\in\{1,2,3,4\}$ in Figure~\ref{fig:ablation}(\subref{fig:ablation-rounds}). The first round updates candidate plans from the loads induced by the initial preferences. Subsequent rounds recompute these loads after the preferences change, allowing plans to respond to how other agents have adjusted their selections. One round performs comparably on the scheduling problems but yields a higher mean performance ratio on MACTP, while the second closes this gap. This difference is consistent with MACTP's longer plan traces capturing multi-step interactions, whereas the scheduling tasks use $H_p=1$. Further rounds provide no consistent improvement, and runtime remains nearly unchanged.

\FloatBarrier

\section{Detailed Algorithms}
\label{app:algorithms}

Algorithm~\ref{alg:pact} summarizes how PACT maps the graph observation at one decision step to a
joint action. The three algorithms below develop this procedure in sequence: candidate plan generation
(Algorithm~\ref{alg:candidates}), coordination on the plan-augmented trace-graph
(Algorithm~\ref{alg:coordination}), and plan selection followed by execution
(Algorithm~\ref{alg:selection}). Together, they complement the corresponding descriptions in
\Cref{sec:candidates}-\Cref{sec:plan-selection} with a more explicit account of the forward
pass.

\begin{algorithm}[h]
\caption{Candidate plan generation for agent $i$ (detail of \Cref{sec:candidates})}
\label{alg:candidates}
\begin{algorithmic}[1]
\STATE \textbf{Input:} {graph observation} $\mathcal{G}_t$; {encodings $h_G$ and $\sigma_i$ from Eq.~\eqref{eq:encode}}; agent $i$; candidate plan count $K$; plan horizon $H_p$
\STATE \textbf{Output:} candidate plans $\hat{p}_i^1, \ldots, \hat{p}_i^K$ and their traces $\mathrm{Tr}(\hat{p}_i^k)$
\STATE {$\hat{x}_{i,0}^k \leftarrow$ current state of agent $i$ from $\mathcal{G}_t$ for every $k$}
\STATE draw $\hat{a}_{i,t}^1, \ldots, \hat{a}_{i,t}^K$ jointly, without replacement, from {$\pi_\theta^{\mathrm{plan}}(\cdot \mid h_G, \sigma_i, \hat{x}_{i,0}^k)$} 
\FOR{$k = 1, \ldots, K$}
    \STATE {$\hat{x}_{i,1}^k \leftarrow$ rollout state of agent $i$ after $\hat{a}_{i,t}^k$}
    \FOR{$j = 1, \ldots, H_p-1$}
        \STATE sample {$\hat{a}_{i,t+j}^k \sim \pi_\theta^{\mathrm{plan}}(\cdot \mid h_G, \sigma_i, \hat{x}_{i,j}^k)$}
        \STATE {$\hat{x}_{i,j+1}^k \leftarrow$ rollout state after $\hat{a}_{i,t+j}^k$}
    \ENDFOR
    \STATE $\hat{p}_i^k \leftarrow (\hat{a}_{i,t}^k, \ldots, \hat{a}_{i,t+H_p-1}^k)$
    \STATE {$\mathrm{Tr}(\hat{p}_i^k) \leftarrow \{(v, d(v,\hat{a}_{i,t+j}^k), j) : d(v,\hat{a}_{i,t+j}^k) > 0\}$} 
\ENDFOR
\RETURN $\{\hat{p}_i^k, \mathrm{Tr}(\hat{p}_i^k)\}_{k=1}^{K}$
\end{algorithmic}
\end{algorithm}

\begin{algorithm}[h]
\caption{Coordination on the plan-augmented trace-graph (detail of \Cref{sec:graph})}
\label{alg:coordination}
\begin{algorithmic}[1]
\STATE \textbf{Input:} candidate plans $\{\hat{p}_i^k, \mathrm{Tr}(\hat{p}_i^k)\}$ for all $i \in \mathcal{M}_t$, $k = 1,\ldots,K$; {coordination rounds $R$}
\STATE \textbf{Output:} {candidate plan} scores $s_{i}^{(R)}$ for all $i \in \mathcal{M}_t$
\STATE $\chi_t \leftarrow \mathcal{G}_t$ with one {candidate plan node} per $\hat{p}_i^k$ and {a trace arc $(\hat{p}_i^k,v)$} for every {$(v,d,j)\in\mathrm{Tr}(\hat{p}_i^k)$} 
\STATE $h^{(0)} \leftarrow$ {$h_G$ for element nodes and an encoding of the node's trace for candidate plan nodes}
\STATE $s_{i}^{(0)}(k) \leftarrow {\eta\,\tanh\big(}f_{\mathrm{score}}(h^{(0)}(\hat{p}_i^k)){\big)}$, \quad $q_{i}^{(0)} \leftarrow {\mathrm{softmax}}(s_{i}^{(0)})$ ({Eq.~\eqref{eq:plan_preference}})
\FOR{$r = 0, \ldots, R-1$}
    \FOR{each element $v$ {on which some trace places demand}}
        \STATE {$L^{(r)}(v) \leftarrow \sum_{i}\sum_{k} q_{i}^{(r)}(k)\, d(v, \hat{p}_i^k)$} (Eq.~\eqref{eq:load})
        \STATE $\tilde{h}^{(r)}(v)
    =
    h^{(r)}(v)
    +
    W_{\mathrm{load}}
    \left[
        \frac{L^{(r)}(v)}{\kappa_t(v)},\;
        \frac{
            \mathrm{ReLU}\!\left(L^{(r)}(v)-(\kappa_t(v)-d_t(v))\right)
        }{
            \kappa_t(v)
        }
    \right],$ (Eq.~\eqref{eq:inject}: {normalized by the capacity $\kappa_t(v)$})
    \ENDFOR
    \STATE $h^{(r+1)} \leftarrow {\mathrm{MPNN}}(\tilde{h}^{(r)}, \chi_t)$ (Eq.~\eqref{eq:mp})
    \STATE $s_{i}^{(r+1)}(k) \leftarrow s_{i}^{(r)}(k) + {\eta\,\tanh\big(}f_{\mathrm{score}}(h^{(r+1)}(\hat{p}_i^k)){\big)}$ (Eq.~\eqref{eq:score})
    \STATE $q_{i}^{(r+1)} \leftarrow {\mathrm{softmax}}(s_{i}^{(r+1)})$ 
\ENDFOR
\RETURN $\{s_{i}^{(R)}\}_{i \in \mathcal{M}_t}$
\end{algorithmic}
\end{algorithm}

\begin{algorithm}[t]
\caption{Plan selection and execution (detail of \Cref{sec:plan-selection})}
\label{alg:selection}
\begin{algorithmic}[1]
\STATE \textbf{Input:} candidate plans $\{\hat{p}_i^k\}$ and {candidate plan} scores $\{s_{i}^{(R)}\}$ for all $i \in \mathcal{M}_t$; selection temperature $\tau$
\STATE \textbf{Output:} executed joint action $\mathbf{a}_t$
\FOR{each agent $i \in \mathcal{M}_t$ {(in parallel)}}
    \STATE $z_i \sim {\mathrm{softmax}}(s_{i}^{(R)}/\tau)$ 
    \STATE {$a_{i,t} \leftarrow \hat{p}_i^{z_i}[0]$} 
\ENDFOR
\STATE $\mathbf{a}_t \leftarrow \{a_{i,t}\}_{i \in \mathcal{M}_t}$
\STATE submit $\mathbf{a}_t$ to the environment, update the state, observation, and cost (\Cref{sec:plan-selection})
\RETURN $\mathbf{a}_t$
\end{algorithmic}
\end{algorithm}

\end{document}